\documentclass[journal]{IEEEtran}

\usepackage{cite}
\usepackage{amsmath,amssymb,amsfonts}
\usepackage{algorithm}
\usepackage{algorithmic}
\usepackage{graphicx}
\usepackage{textcomp}

\usepackage{bm}
\usepackage{xcolor}
\usepackage{pict2e}

\newsavebox{\ORCIDlogo}
\savebox{\ORCIDlogo}{%
	\setlength{\unitlength}{\dimexpr 0.75em/256\relax}%
	\begin{picture}(256,256)%
		\color[HTML]{A6CE39}\put(128,128){\circle*{256}}%
		\color{white}%
		\put(78.6,199.2){\circle*{20}}%
		\moveto(70.9,176,9)\lineto(86.3,176,9)\lineto(86.3,69.8)\lineto(70.9,69.8)%
		\closepath\fillpath%
		\moveto(108.9,176.9)\lineto(150.5,176.9)%
		\curveto(190.1,176.9)(207.5,148.6)(207.5 ,123.3)%
		\curveto(207.5,95,8)(186,69.7)(150.7,69.7)%
		\lineto(108.9,69.7)%
		\closepath\fillpath%
		\color[HTML]{A6CE39}%
		\moveto(124.3,83.6)\lineto(148.8,83.6)%
		\curveto(183.7,83.6)(191.7,110.1)(191.7,123.3)%
		\curveto(191.7,144.8)(178,163)(148,163)%
		\lineto(124.3,163)%
		\closepath\fillpath%
	\end{picture}%
}

\newcommand\orcidicon[1]{\href{https://orcid.org/#1}{\usebox{\ORCIDlogo}}}

\usepackage[
colorlinks=true,
linkcolor=blue,
urlcolor=blue,
citecolor=blue
]{hyperref}

\begin{document}
%
% paper title
% Titles are generally capitalized except for words such as a, an, and, as,
% at, but, by, for, in, nor, of, on, or, the, to and up, which are usually
% not capitalized unless they are the first or last word of the title.
% Linebreaks \\ can be used within to get better formatting as desired.
% Do not put math or special symbols in the title.
\title{Adaptive Training of Grid-Dependent Physics-Informed Kolmogorov-Arnold Networks}
%
%
% author names and IEEE memberships
% note positions of commas and nonbreaking spaces ( ~ ) LaTeX will not break
% a structure at a ~ so this keeps an author's name from being broken across
% two lines.
% use \thanks{} to gain access to the first footnote area
% a separate \thanks must be used for each paragraph as LaTeX2e's \thanks
% was not built to handle multiple paragraphs
%

\author{Spyros Rigas$^{\orcidicon{0009-0009-2352-8709}}$, Michalis Papachristou$^{\orcidicon{0000-0001-5650-4206}}$, Theofilos Papadopoulos$^{\orcidicon{0000-0002-8534-785X}}$, Fotios Anagnostopoulos$^{\orcidicon{0000-0003-0176-4144}}$ and Georgios Alexandridis$^{\orcidicon{0000-0002-3611-8292}}$% <-this % stops a space
\thanks{Spyros Rigas and Georgios Alexandridis are with the Department of Digital Industry Technologies, School of Science, National and Kapodistrian University of Athens, 34400 Greece (e-mail: spyrigas@uoa.gr; gealexandri@uoa.gr).}% <-this % stops a space
\thanks{Michalis Papachristou is with the Department of Physics, School of Science, National and Kapodistrian University of Athens, 15784 Greece (e-mail: mixpap@phys.uoa.gr).}% <-this % stops a space
\thanks{Theofilos Papadopoulos is with the School of Electrical \& Computer Engineering, National Technical University of Athens, 15780 Greece (e-mail: teopap@mail.ntua.gr).}% <-this % stops a space
\thanks{Fotios Anagnostopoulos is with the Department of Informatics \& Telecommunications, University of the Peloponnese, 22131 Greece (e-mail: fotisanagn@uop.gr).}% <-this % stops a space
}

\maketitle

% As a general rule, do not put math, special symbols or citations
% in the abstract or keywords.
\begin{abstract}
Physics-Informed Neural Networks (PINNs) have emerged as a robust framework for solving Partial Differential Equations (PDEs) by approximating their solutions via neural networks and imposing physics-based constraints on the loss function. Traditionally, Multilayer Perceptrons (MLPs) have been the neural network of choice, with significant progress made in optimizing their training. Recently, Kolmogorov-Arnold Networks (KANs) were introduced as a viable alternative, with the potential of offering better interpretability and efficiency while requiring fewer parameters. In this paper, we present a fast JAX-based implementation of grid-dependent Physics-Informed Kolmogorov-Arnold Networks (PIKANs) for solving PDEs, achieving up to 84 times faster training times than the original KAN implementation. We propose an adaptive training scheme for PIKANs, introducing an adaptive state transition technique to avoid loss function peaks between grid extensions, and a methodology for designing PIKANs with alternative basis functions. Through comparative experiments, we demonstrate that the adaptive features significantly enhance solution accuracy, decreasing the L$^2$ error relative to the reference solution by up to 43.02\%. For the studied PDEs, our methodology approaches or surpasses the results obtained from architectures that utilize up to 8.5 times more parameters, highlighting the potential of adaptive, grid-dependent PIKANs as a superior alternative in scientific and engineering applications.
\end{abstract}

% Note that keywords are not normally used for peerreview papers.

\section{Introduction} \label{sec:introduction}

The ever-present need to address problems related to differential equations for modeling complex systems, combined with the exponential progress in Deep Learning (DL), has led to the emergence of a new paradigm: Physics-Informed Neural Networks (PINNs) \cite{pinns1,pinns2,pinns3}. PINNs provide a framework for solving partial differential equations (PDEs) by transforming them into a loss function optimization problem. Specifically, PINNs employ a deep neural network to represent the solution of a PDE within a given domain. Incorporating experimental data is essential for inverse problems, however in the case of forward problems it is sufficient to sample points within the domain, known as collocation points, and train the network in an unsupervised manner by encoding the underlying equation and boundary conditions as terms of the loss function. This approach offers several advantages over traditional methods, including the ability to provide mesh-free solutions, generate solutions with tractable analytical gradients and jointly solve forward and inverse problems within the same framework.

Despite their relatively recent introduction to the mainstream DL literature, PINNs have already demonstrated significant potential and versatility in various scientific and engineering applications. Notable examples include the simulation of fluid flows as described by the Euler \cite{euler} or Navier-Stokes equations \cite{navier1,navier2}, applications spanning from medicine \cite{med1, med2} to electric power systems \cite{powersystems1, powersystems2}, and even domain-specific problems in physics, such as modeling astrophysical shocks \cite{vlahakis} or calculating eigenfunctions and eigenvalues for quantum mechanical systems \cite{mattheakis}. Nonetheless, PINNs are not without shortcomings, which can stem from the framework itself, such as weighting imbalances in the loss function's terms \cite{lossannealing}, or the underlying DL architecture, for instance, the issue of spectral bias inherent in Multilayer Perceptrons (MLPs) with Rectified Linear Unit (ReLU) activations \cite{spectralbias}. To mitigate such issues, numerous studies have focused on improving the performance of PINNs by proposing alternative architectures beyond the simple, yet prevalent MLP \cite{lossannealing,archs1, archs2, moseley}, or by employing adaptive training strategies. The latter include, but are not limited to, adaptive re-weighting of the loss function's terms \cite{lossannealing,loss-colloc, ntk, lbpinns, relobralo, rba}, embedding gradient information of the PDE residuals in the loss function \cite{gpinns, sobolev}, adaptive re-sampling of collocation points \cite{loss-colloc, colloc1, colloc2}, and utilizing adaptive activation functions \cite{activations1, activations2}.

Inspired by the Kolmogorov-Arnold representation theorem, a novel architecture called Kolmogorov-Arnold Networks (KANs) was recently proposed as an alternative to MLPs \cite{pykan}. Unlike in the case of MLPs, the computational graph of KANs involves learnable activation functions at its edges and a sum operation on its nodes, aiming to achieve higher accuracy and interpretability, while utilizing a considerably smaller number of parameters. Due to their potential, KANs have already been successfully adopted for tasks like time-series analysis \cite{timekan1, timekan2, timekan3}, image recognition \cite{satellitekan,compviskan}, image segmentation \cite{ukan} and human activity recognition \cite{harkan}. However, their most promising application lies in addressing complex scientific and engineering problems that can be formulated as (symbolic) regression or PDE solving \cite{pykan}. In this respect, the term Physics-Informed Kolmogorov Arnold Networks (PIKANs) has been coined to refer to PINNs that are based on KANS for their underlying architecture \cite{karniadakisreview}.

Naturally, KANs are ideal candidates for the PINN framework, as they correspond to an innovative architecture with inherent adaptability in its activation functions. Preliminary studies on PIKANs have shown results comparable \cite{karniadakisreview} or superior \cite{pykan, ropinn} to those of MLPs. Nevertheless, this adaptability is also the cause of a significant drawback; the training of KANs, and by extension PIKANs, is computationally expensive, as a consequence of using learnable B-Splines as activation functions. To address this issue, following studies have proposed more computationally efficient activation functions, such as radial basis functions \cite{rbfkan}, Chebyshev polynomials \cite{chebykan}, wavelets \cite{wavkan}, and ReLU-based functions \cite{relukan}. For the latter, a speedup factor between 5 and 20 in training time has been reported compared to the original KAN implementation.

Albeit successful in increasing their computational efficiency, a common pattern in all the aforementioned alternatives is that they remove the dependence of PIKANs on the grid. The grid is the set of points used to construct the basis functions, whose linear combinations correspond to the learnable activations. Chebyshev (and other Jacobi) polynomials are inherently grid-independent, while the implementations involving radial basis functions, wavelets, and ReLU-based functions focus on uniform or static grids, which lead to fixed basis functions. Herein, we argue that this may adversely affect the training of PIKANs, resulting in slower convergence and potentially less accurate solutions. Additionally, these implementations fail to leverage one of the key features of KANs which has been reported to significantly enhance their accuracy \cite{pykan}: the ability to dynamically extend the grid during training, thereby obtaining progressively more fine-grained basis functions.

%In this study, we aim to advance the current research on PIKANs by proposing practices that can accelerate their training process without making compromises on their accuracy.

The contributions of the current work can be summarized in the following points:

\begin{itemize}
	\item{The introduction of a new, open-source computational framework for KANs, developed in JAX \cite{jax} and Flax \cite{flax} to significantly accelerate the training of KANs.}
	\item{The study of an open issue involving abrupt jumps in the loss function’s values after grid extensions and the introduction of an adaptive transition method to address it and further reduce the model's training loss.}
	\item{The adaptation of loss re-weighting and collocation re-sampling schemes to a grid-dependent framework for training PIKANs with relatively small architectures.}
	\item{The introduction of the concepts of staticity and full grid adaptivity in the design of (PI)KANs with alternative basis functions, emphasizing the importance of preserving their dependency on the grid to enable more adaptive training.}
\end{itemize}

The remainder of this paper is structured as follows; in Section \ref{sec:framework} we formally review the PIKAN framework and introduce our JAX-based implementation, including benchmarks compared to the original KAN implementation. Section \ref{sec:training} presents the proposed adaptive training schemes, aiming to form the foundation for future standardized end-to-end pipelines in PIKAN training, inspired by \cite{pinnpipeline}. Additionally, we provide a solution which involves regular grid adaptations and a simple linear interpolation of the optimizer's parameters during training, to solve the issue of the loss function's sharp peaks after grid extensions. A thorough discussion on KAN basis functions follows in Section \ref{sec:grid}, emphasizing the effect of grid dependency on PIKAN accuracy. Using ReLU-KANs as a case study, we present a method to make the basis functions fully adaptive to the grid. Finally, in Section \ref{sec:conclusion}, we summarize the main results of our work and propose avenues for future research on the field of PIKANs.

\section{PIKAN Framework} \label{sec:framework}

Before discussing the technical details of the JAX-based implementation developed to accelerate the training of PIKANs, we provide an overview of their constituents.

\subsection{PINN Problem Formulation}

Consider a differential equation of the generic form

\begin{align}
	\mathcal{F}_\lambda^{}\left(u\left(x\right)\right) &= f\left(x\right), ~x \in \Omega \nonumber, \\
	\mathcal{B}_\lambda^{k}\left(u\left(x\right)\right) &= b_{}^{k}\left(x\right), ~x \in \Gamma^k \subseteq \partial \Omega, \label{eq1}
\end{align}

\noindent where $x$ is the coordinate vector, $u$ is the equation's solution, $\lambda$ corresponds to a set of model parameters, $\mathcal{F}$ is a nonlinear differential operator and $\left\{\mathcal{B}_{}^k\right\}_{k=1}^{n_\Gamma}$ is a set of $n_\Gamma$ boundary condition operators. For time-dependent problems, the temporal coordinate is integrated in $x$, therefore the problem's initial conditions are considered as a special case of boundary conditions in $\Omega$. The main intuition behind PINNs is to approximate the exact solution with a neural network, $u\left(x;\theta\right)$, parameterized by a set of parameters, $\theta$, and minimize the loss function, $\mathcal{L}\left(\theta\right)$, which encapsulates the differential equation and boundary conditions as distinct residual terms.

The training of PINNs generally involves three different types of datasets. The first, $\mathcal{D}_f$, is a dataset consisting of $N_f$ collocation points which lie in the interior of $\Omega$ and are used to enforce the physics captured by the differential equation. The second, $\left\{\mathcal{D}_b^k\right\}_{k=1}^{n_\Gamma}$, is a series of datasets containing $N_b^k$ collocation points each, which are used to enforce the boundary conditions on $\Gamma^k$. Finally, $\mathcal{D}_d$ is a dataset comprising tuples of experimental data coordinates and measured values which add a supervised component to the network's training. The latter is necessary only when $\lambda$ are unknown, corresponding to the so-called inverse problem of differential equations. Herein, we only study forward problems, meaning that all $\lambda$ are considered known and $\mathcal{D}_d$ is null. Therefore the loss function, often referred to as the physics-informed loss, assumes the form

\begin{equation}
	\mathcal{L}\left(\theta\right) = w_f\mathcal{L}_f\left(\theta\right) + \sum_{k=1}^{n_\Gamma}{w_b^k\mathcal{L}_b^k\left(\theta\right)}, \label{eq2}
\end{equation}

\noindent where $w_f$ and $\left\{w_b^k\right\}_{k=1}^{n_\Gamma}$ are individual weights of the loss function's terms and

\begin{align}
	\mathcal{L}_f\left(\theta\right) &= \frac{1}{N_f}\sum_{i=1}^{N_f}{\lVert \mathcal{F}\left(u\left(x_i;\theta\right)\right) - f\left(x_i\right) \rVert^2}, \label{eq3} \\
	\mathcal{L}_b^k\left(\theta\right) &= \frac{1}{N_b^k}\sum_{i=1}^{N_b^k}{\lVert \mathcal{B}^k\left(u\left(x_i;\theta\right)\right) - b^k\left(x_i\right) \rVert^2}, \label{eq4}
\end{align}

\noindent with $\lVert \cdot \rVert$ denoting the L$^2$ norm. Derivatives of $u\left(x;\theta\right)$ of arbitrary order and with respect to any coordinate that may appear in \eqref{eq3} or \eqref{eq4} are computed using automatic differentiation (AD) \cite{ad}.

\subsection{Kolmogorov-Arnold Networks}

In the vast majority of problems involving PINNs, $u\left(x;\theta\right)$ corresponds to a MLP with fixed number of linear layers, each followed by a non-linear activation. Denoting $\theta = \left\{W^{(l)},b^{(l)}\right\}_{l=1}^{L}$, where $W^{(l)}$ and $b^{(l)}$ is the weight matrix and bias of the $l$-th layer, respectively, an MLP with $L$ layers and activation functions $\sigma$ can be written as

\begin{equation}
	u\left(x;\theta\right) = \left[\Phi^{(L)}\circ \dots \circ \Phi^{(1)}\right]\left(x\right), \label{eq5}
\end{equation}

\noindent where $\circ$ denotes successive application of $\Phi^{(l)}$, and

\begin{equation}
	\Phi^{(l)}\left(x^{(l)}\right) = \sigma\left(W^{(l)}x^{(l)} + b^{(l)}\right). \label{eq6}
\end{equation}

\noindent In the case of KANs, \eqref{eq5} still holds, however the non-linear activation on the linear combination of the layer's inputs is replaced by

\begin{equation}
	\Phi^{(l)}\left(x^{(l)}\right) = 	
	\begin{pmatrix}
		\phi_{l,1,1}(\cdot) & \cdots & \phi_{l,1,n_l}(\cdot) \\
		\phi_{l,2,1}(\cdot) & \cdots & \phi_{l,2,n_l}(\cdot) \\
		\vdots & \ddots & \vdots \\
		\phi_{l,n_{l+1},1}(\cdot) & \cdots & \phi_{l,n_{l+1},n_l}(\cdot)
	\end{pmatrix} x^{(l)}, \label{eq7}
\end{equation}

\noindent where $n_l$ is the number of input nodes for the $l$-th layer and $\phi_{l,i,j}$ is the $l$-th layer's univariate activation function, connecting its $i$-th input node to its $j$-th output node in the network's computational graph. Obviously, a layer's output node coincides with the next layer's input node. Essentially, in KANs, the concept of the layer does not revolve around a set of nodes, but instead refers to the edges, which is where the activation functions lie. For a layer with $n_l$ input nodes and $n_{l+1}$ output nodes, the number of univariate activation functions is the product $n_l\cdot n_{l+1}$. To refer to the architecture of a KAN, it suffices to refer to its shape as an array of integers, where two consecutive numbers correspond to a layer's input and output nodes. For instance, the architecture of a $L$-layered KAN is written as $\left[n_0, n_1, \dots, n_L\right]$.

\begin{figure}[!t]
	\centering
	\includegraphics[width=\linewidth]{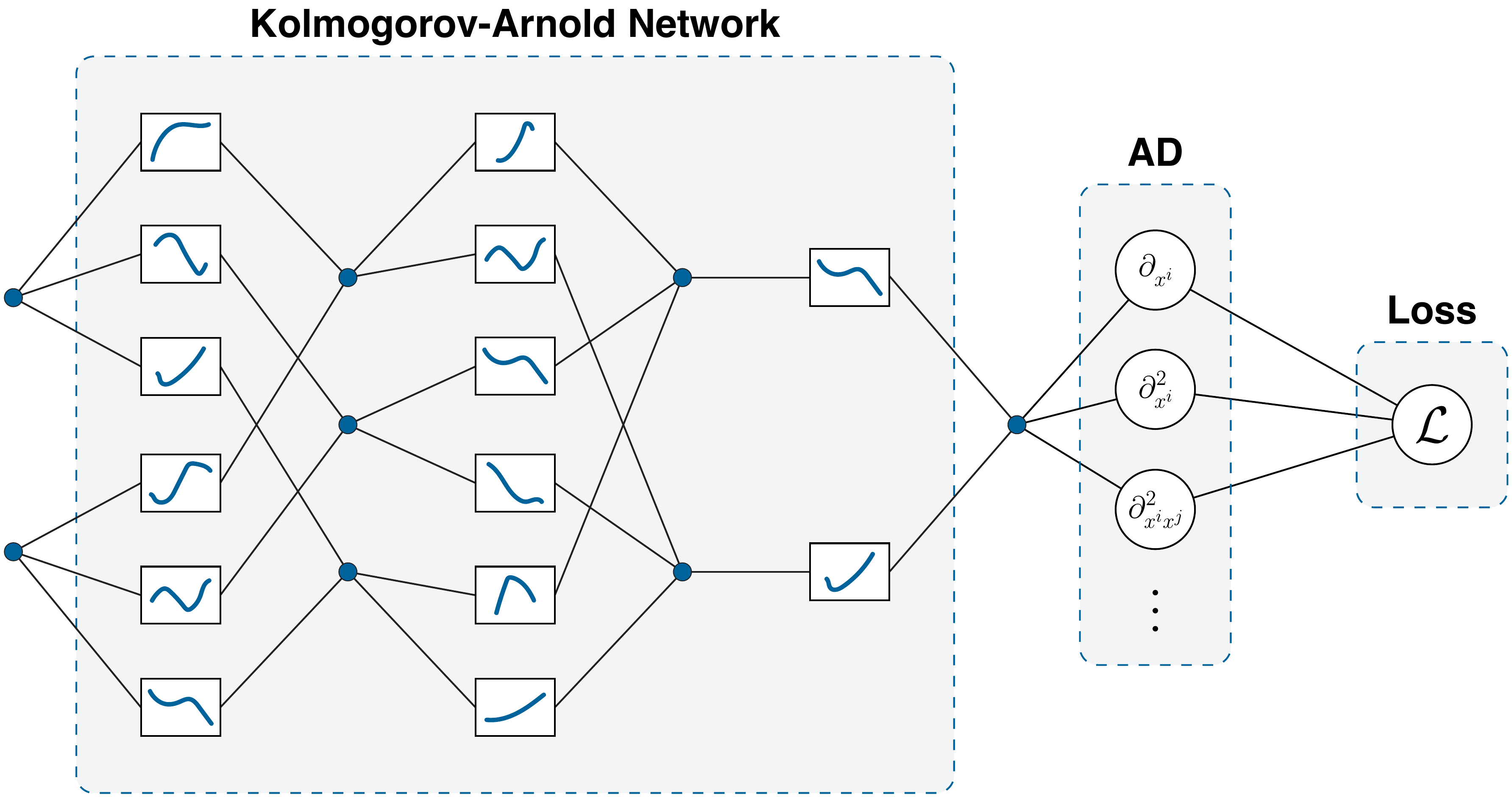}
	\caption{Schematic representation of a PIKAN with an underlying [2,3,2,1] KAN architecture.}
	\label{pikan}
\end{figure}

In the original implementation of KANs \cite{pykan}, referred to as ``vanilla'' KANs in \cite{karniadakisreview}, the univariate activation functions are given by:

\begin{equation}
	\phi\left(x\right) = c_rr\left(x\right) + c_BB\left(x\right), \label{eq8}
\end{equation}

\noindent where

\begin{equation}
	r\left(x\right) = \frac{x}{1+ \exp\left(-x\right)} \label{eq9}
\end{equation}

\noindent is a residual-like activation and

\begin{equation}
	B\left(x\right) = \sum_{i=1}^{G+k}{c_i B_i\left(x\right)} \label{eq10}
\end{equation}

\noindent is a B-Spline activation of order $k$, defined on a grid (knot vector) with $G$ intervals. For a given grid and value of $k$, a set of spline basis functions $\left\{B_i\right\}_{i=1}^{G+k}$ is uniquely defined, constituting the activation functions of \eqref{eq8} grid-dependent by design. The parameters $c_r$, $c_B$ and $\left\{c_i\right\}_{i=1}^{G+k}$ are trainable, which is why in KANs the activation functions are not fixed, unlike in MLPs. Based on these, extending the framework of PINNs to PIKANs is straightforward: simply replace the expression for $\Phi^{(l)}$ from \eqref{eq7} in \eqref{eq5} and use the obtained $u\left(x;\theta\right)$ in \eqref{eq3}, \eqref{eq4}. A schematic representation of an example PIKAN along with its underlying KAN's computation graph can be seen in Fig. \ref{pikan}.

Importantly, the introduction of grid-dependent trainable activations in KANs is what gives rise to a major advantage of MLPs over them, which is related to their complexity. In particular, if one considers an MLP with $L$ layers, each having $N$ neurons, then the number of trainable parameters is $\mathcal{O}\left(N^2L\right)$. In contrast, for a $L$-layered $[N,N,\dots,N]$ KAN the situation is quite different, as the number of trainable parameters is $\mathcal{O}\left(N^2\left(G+k\right)L\right)$. Consequently, to make the training times of the two network types comparable, one may attempt to train narrower KANs (smaller values of $N$), as they have been found to perform comparably to or even outperform MLPs with wider architectures \cite{pykan}. In addition, the training time of KANs can be further decreased by dynamically increasing the grid size, $G$, rather than training the network with a fixed, large value of $G$. This process is referred to as grid extension and involves initiating training with a small value of $G$ and progressively increasing it as a type of fine-graining process.

Apart from grid extension, another technique for training KANs is referred to as grid adaptation and involves regularly altering the grid's points based on the values of the corresponding activation function. To perform a grid adaptation, the activation function's inputs are sorted, and the minimum and maximum values are chosen as the grid's ends. Then, a uniform grid, $\mathcal{G}_u$, is constructed by placing equidistant points between the grid's ends. Simultaneously, a fully adaptive grid, $\mathcal{G}_a$, is constructed by choosing the interior points based on the distribution of the activation function's inputs. The final grid, $\mathcal{G}$, is then given as a linear mixing of the two:

\begin{equation}
	\mathcal{G} = g_e\mathcal{G}_u + \left(1-g_e\right)\mathcal{G}_a, \label{eq11}
\end{equation}

\noindent where $0 \leq g_e \leq 1$ is the hyper-parameter that controls whether the grid tends to be more uniform ($g_e > 0.5$) or more adaptive to the inputs ($g_e < 0.5$). When combined, the processes of grid extension and grid adaptation can be jointly referred to as a grid update.

\subsection{JAX Implementation}

\begin{figure*}[!t]
	\centering
	\includegraphics[width=\linewidth]{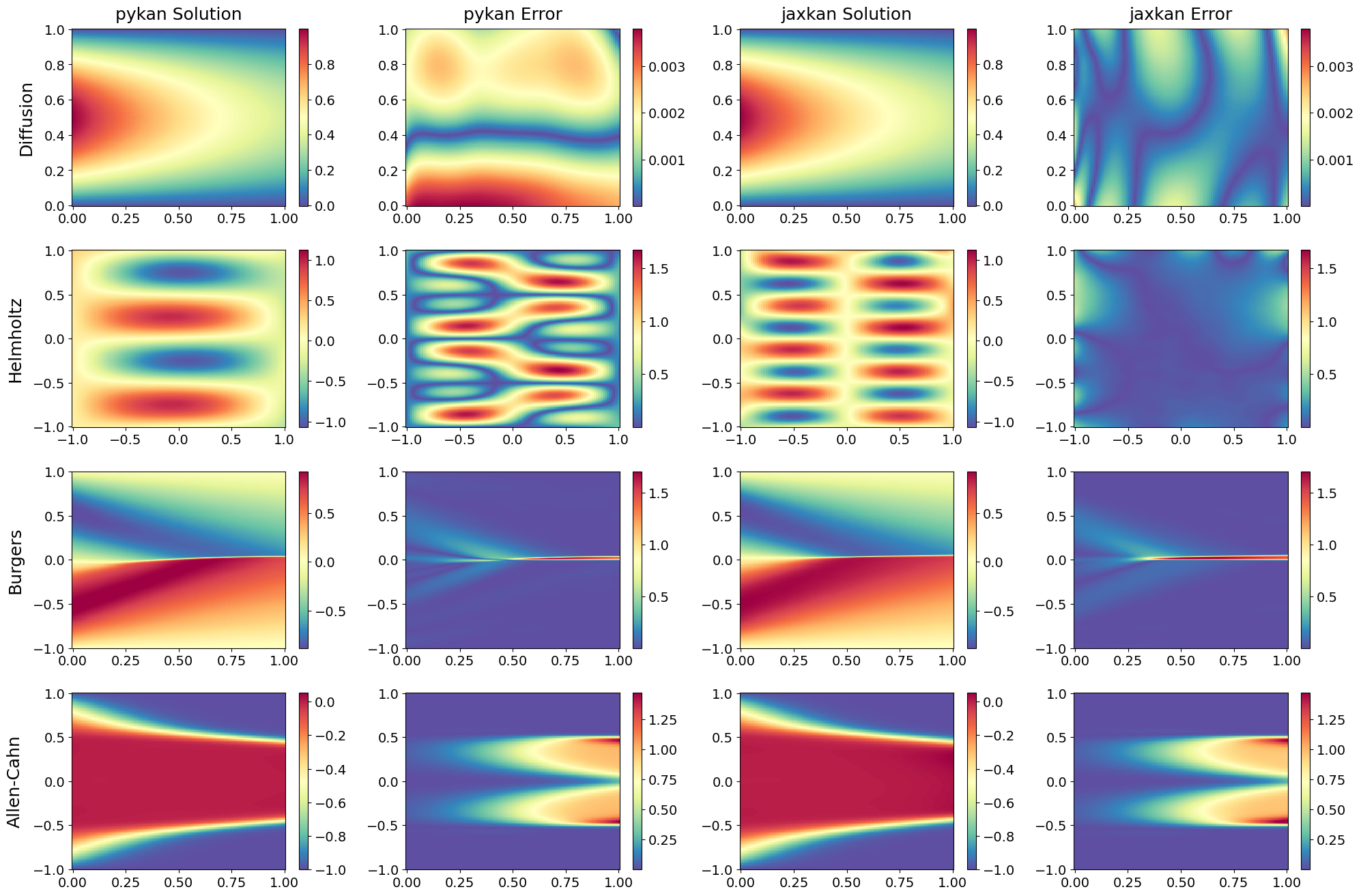}
	\caption{PIKAN results for the Diffusion equation (first row), the Helmholtz equation (second row), Burgers' equation (third row), and the Allen–Cahn equation (fourth row). The first/third and second/fourth columns correspond to the solution obtained via pykan/jaxkan and its absolute error compared to the reference solution, respectively. In each row, the solutions/errors share the same colorbars.}
	\label{baseline-pikan}
\end{figure*}

JAX is a numerical computing library that combines the flexibility, just-in-time compilation and parallelization capabilities of high-performance frameworks, with the GPU/TPU acceleration and AD features of DL frameworks like PyTorch \cite{pytorch} and TensorFlow \cite{tensorflow}. Thanks to these, JAX has become increasingly popular for developing PINN-related code \cite{moseley,pinnpipeline}, offering significant performance improvements in both training speed and computational efficiency. Leveraging these advantages, we developed a JAX-based implementation of KANs called \texttt{jaxKAN}, with the aim of accelerating the training process of PIKANs. For this reason, features relevant to symbolic regression, available in the original \texttt{pykan} implementation \cite{pykan}, were not included. Instead, we focused on methods and utilities relevant specifically to PIKANs. Our implementation is available as a Python package in PyPI \cite{jaxkan} and has already been used for research on PIKANs \cite{fbkans}.

The \texttt{jaxKAN} package is organized into three main modules. The \texttt{models} module includes the \texttt{KANLayer} and \texttt{KAN} classes, which inherit from Flax's \texttt{linen} module and correspond to the vanilla KAN implementation. This module is designed to also accommodate alternative implementations of KANs. A notable example of this are ReLU-KANs, which were recently proposed as a more efficient alternative to B-Spline-based KANs, since ReLU-based operations can be fully parallelized and are therefore more hardware friendly - especially for GPUs/TPUs. The \texttt{bases} module includes functions related to the basis functions used to construct the learnable activations. For vanilla KANs, these are B-Splines, calculated using an iterative version of the Cox-de Boor recursion formula \cite{deboor,cox}. Finally, the \texttt{utils} module contains utility functions pertinent to the functionality of KANs, such as a parallel implementation of the least squares algorithm (used to re-initialize the $\left\{c_i\right\}_{i=1}^{G+k}$ coefficients after grid extensions) based on JAX's \texttt{vmap}. Additionally, it includes methods required for the adaptive training of PIKANs, as further explained in Section \ref{sec:training}.

In addition to using a different framework and including implementations beyond the vanilla KAN, \texttt{jaxKAN} has two significant differences compared to \texttt{pykan}. First, we developed the grid update as an internal model method that can simultaneously adapt the grid to the data and extend its size. As demonstrated in subsequent experiments, the grid has a considerable impact on the performance of KANs, therefore fine-tuning the hyperparameters related to its update is crucial when training PIKANs. The second major difference is our choice of optimizer. Instead of the resource-intensive LBFGS \cite{lbfgs}, we opted for the Adam optimizer \cite{adam} implemented through the Optax optimization framework \cite{optax}. Notably, while more recent implementations of KANs have aimed to enhance training efficiency by proposing alternative KAN layer structures \cite{efficientkan}, we chose not to use them due to their inability to produce the regularization terms presented in the original paper.

\begin{table}[!t]
	\caption{\textbf{Average GPU training time per epoch and standard error, measured on NVIDIA's GeForce RTX-4070 12GB.}}
	\centering
	\setlength{\tabcolsep}{12pt}
	\begin{tabular}{|c|c|c|}
		\hline
		\rule{0pt}{3ex} PDE & \texttt{pykan} (ms) & \texttt{jaxKAN} (ms) \\ [0.5ex]
		\hline
		\rule{0pt}{3ex} Diffusion Equation & $189 \pm 2$ & $\mathbf{2.6} \boldsymbol{\pm} \mathbf{0.1}$ \\
		\rule{0pt}{3ex} Helmholtz Equation & $237 \pm 2 $ & $\mathbf{2.8} \boldsymbol{\pm} \mathbf{0.2}$ \\
		\rule{0pt}{3ex} Burgers' Equation & $231 \pm 3 $ & $\mathbf{3.1} \boldsymbol{\pm} \mathbf{0.2}$ \\
		\rule{0pt}{3ex} Allen-Cahn Equation & $194 \pm 5 $ & $\mathbf{2.9} \boldsymbol{\pm} \mathbf{0.1}$ \\ [0.5ex]
		\hline
	\end{tabular}
	\label{times}
\end{table}

Before reviewing our adaptive training framework, we provide in Fig. \ref{baseline-pikan} the results of vanilla PIKANs which were trained for $5\cdot 10^{4}$ epochs to solve four different PDEs: the diffusion equation (first row), the Helmholtz equation (second row), Burgers' equation (third row), and the Allen–Cahn equation (fourth row). These PDEs are among the most commonly selected in PINN-related problems, as each of them presents different challenges: the diffusion equation is the simplest benchmark used merely as a sanity check; the Helmholtz equation is special in that the residuals of the PDE and the boundary conditions usually differ by orders of magnitude; Burgers' equation has a notable discontinuity for $x=0$, which may be hard for a model to capture; and vanilla PINNs tend to struggle in solving the Allen-Cahn equation without some type of adaptive technique implemented for their training \cite{causaltrain}. In addition to these, the former two have closed-form, analytical solutions, where KANs may have an advantage over other architectures \cite{pykan}, while the latter two do not. Each of these PDEs is presented in Appendix \ref{appA}, including the reference solutions required to calculate the absolute errors shown in the second and fourth columns of Fig. \ref{baseline-pikan}, as well as details regarding the sampling of collocation points.

The PIKANs were implemented in both \texttt{jaxKAN} and \texttt{pykan} with identical hyper-parameters and architectures: in every case, $k=3$ was chosen for the spline basis functions' order, the underlying KAN's architecture was $[2,6,6,1]$ and a grid of $G = 3$ intervals was used. The weights of the loss function's terms were all set equal to $1.0$, and a constant value of $10^{-3}$ was selected for the learning rate. With the exception of the Helmholtz equation, where the absolute error for \texttt{jaxKAN} is smaller than that for \texttt{pykan} by an order of magnitude, the two implementations produce similar results, both quantitatively and qualitatively: the diffusion equation's solution is obtained with relatively good accuracy in both implementations; Burgers' equation appears to be well approximated in all regions except for $x = 0$, where the well-known discontinuity exists; the approximated solution for the Allen-Cahn equation is visibly different from the reference solution in both cases.

Nonetheless, a significant discrepancy between the two implementations arises when comparing the PIKANs' training times. As shown in Table \ref{times}, our framework is two orders of magnitude faster than \texttt{pykan}. Importantly, the GPU times reported for \texttt{jaxKAN} include the large latencies incurred by the compilation of some functions. This indicates that, if the same experiments were performed with pre-compiled functions, the gap between the benchmarks for the two implementations would be even wider. To avoid biased comparisons, the \texttt{pykan} code used to train PIKANs was taken from its official repository and was even optimized to an extent, for example by removing function declarations within the training loop, or by calculating second-order derivatives without computing the corresponding Hessian matrix.

\section{Adaptive PIKAN Training} \label{sec:training}

In order to surpass the results shown in Fig. \ref{baseline-pikan} and obtain more accurate solutions for Burgers' and the Allen-Cahn equations, one may increase the number of training epochs and use a deeper and wider PIKAN architecture. While this ``brute-force'' approach has been extensively used in the field of deep learning with proven results, we advocate for utilizing adaptive techniques that can lead to comparably good results with smaller architectures and shorter computation times. Besides, in order for KANs to be competitive against MLPs, they must be able to produce on-par or better results while utilizing an equal number of parameters, or even fewer. A series of such techniques are presented in this Section. It is noted that, due to their significantly faster training times, PIKANs implemented using our \texttt{jaxKAN} framework were used for all subsequent experiments. Additionally, for all trained PIKANs, the order of the spline basis functions was set to $k = 3$, in order to further minimize training times.

\subsection{State transition after extension}

The first technique introduced is unique to PIKANs and pertains to the grid extensions that can be performed during the training of KANs. An issue with this process, which is present in most published works and has been noted by \cite{wavkan}, is the fact that the loss function's values tend to experience sharp increases immediately after it is performed. In fact, this phenomenon is one of the reasons why the authors introducing Wav-KANs argued against using B-Splines as activation functions. In this work, we identify the main cause of this effect not to be connected to the use of B-Splines themselves, but rather to the grid's adaptation that occurs simultaneously with its extension, along with the re-initialization of the optimizer's state which follows the grid's update.

\begin{figure}[!t]
	\centering
	\includegraphics[width=\linewidth]{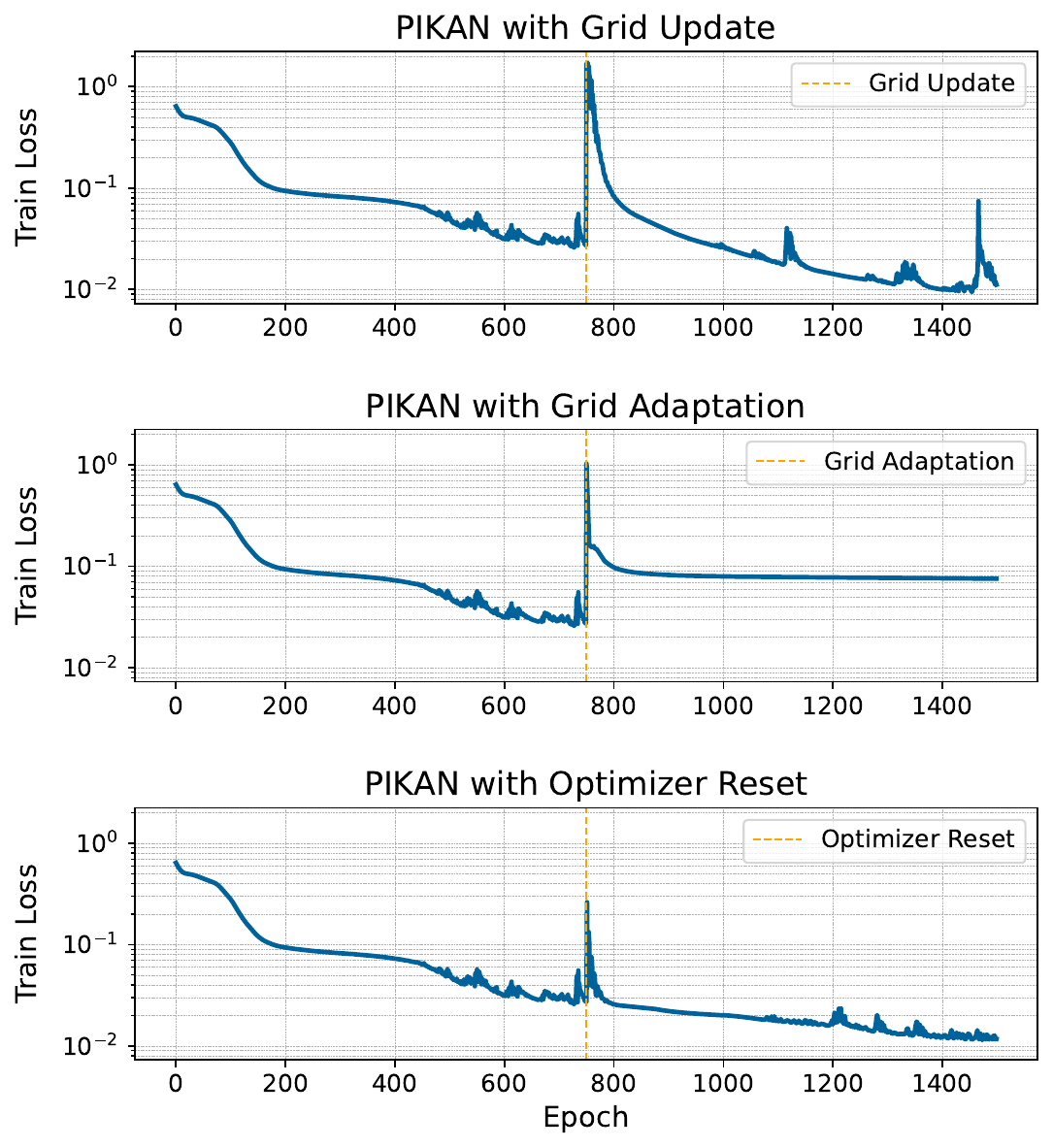}
	\caption{Introducing a grid update (top), grid adaptation (middle) and optimizer reset (bottom) process during the training of a PIKAN.}
	\label{state-upd}
\end{figure}

To showcase this, we trained three identical $[2, 8, 8, 1]$ PIKANs for Burgers' equation for 1500 epochs with a learning rate of $3\cdot 10^{-3}$. For the first PIKAN, the loss function's values of which are shown in the top graph of Fig. \ref{state-upd}, we started from a grid with size $G=4$ and performed a grid update with $G^\prime = 15$ during the 750th training epoch. Following this, the optimizer's state was reset, so that training could continue for the new network parameters. The sharp increase in the loss function's value can be clearly observed as an abrupt discontinuity during the epoch of the grid's update and the subsequent reset of the optimizer's state. To isolate the contribution of the grid's adaptation to this effect, only an adaptation and no extension or state reset was performed during the 750th training epoch of the second PIKAN. The result for the loss function's values can be seen in the middle graph of Fig. \ref{state-upd}, where a similar peak - albeit with lower amplitude - can be observed. For the third PIKAN, we only reset the optimizer's state during the 750th training epoch, without updating the grid at all. A peak can again be observed in the corresponding loss function's values, depicted in the bottom graph of Fig. \ref{state-upd}, although its amplitude is even lower.

To interpret the grid adaptation's contribution, one must note that \eqref{eq11} describes a grid reconstruction that alters the B-spline basis functions, which are by definition grid-dependent. During the training of a network for a number of epochs without grid adaptations, each layer's nodes are attempting to converge to values for which the network's loss function approaches a minimum. If a grid adaptation occurs after this convergence has started taking place, the values of the nodes undergo an abrupt change which in turn induces high gradients and by extension sudden increases for the loss function's values. As far as the re-initialization of the optimizer's state is concerned, it should be a contributing factor for such sharp loss increases in the case of any optimizer with an internal state that handles the scaling of the gradients involved in updating the model's parameters. Adam is a typical example, where gradient scaling is based on estimates of the first and second-order moments of the gradients, using exponential moving averages. Each time Adam is re-initialized, the values for the gradients' first and second moments are reset to zero, essentially erasing information of the gradients' evolution and restarting the training process, albeit with better guesses for the parameters' values.

Based on these findings, we propose an adaptive state transition to remedy this effect, which involves three techniques that have to be incorporated in the training of PIKANs. Firstly, by performing periodic grid adaptations between consecutive grid extensions, one ensures that the variation of the nodes' values occurs progressively throughout training and not abruptly once per extension. Secondly, reducing the learning rate immediately after the grid's update helps in minimizing the contribution of very large gradients in the loss function's value. Finally, in order to avoid the ``hard reset'' of the training process due to the optimizer's state re-initialization after the grid's extension, we propose an optimizer state update described by Algorithm \ref{alg:state_transition}, where $t$ corresponds to the training epoch and $\mu$, $\nu$ are the first and second-order moments of the parameters' gradients, respectively.

\begin{algorithm}[H]
	\caption{Adaptive Optimizer State Transition}
	\label{alg:state_transition}
	\begin{algorithmic}[1]
		\renewcommand{\algorithmicrequire}{\textbf{Input:}}
		\renewcommand{\algorithmicensure}{\textbf{Output:}}
		
		\REQUIRE Old Adam state $\left(t, \mu, \nu\right)$, new grid size $G^\prime \in \mathbb{N}$
		\ENSURE New Adam state $\left(t^\prime, \mu^\prime, \nu^\prime\right)$
		
		\STATE Copy state step:
		\STATE \quad $t^\prime \gets t$
		
		\STATE Copy all moments for $c_r$ and $c_B$:
		\STATE \quad $\mu^\prime\left(c_r,c_B\right) \gets \mu\left(c_r,c_B\right)$
		\STATE \quad $\nu^\prime\left(c_r,c_B\right) \gets \nu\left(c_r,c_B\right)$
		
		\STATE Interpolate new moments for $\left\{c_i\right\}_{i=1}^{G^\prime+k}$ per function $\phi$ per layer $l$:
		
		\FOR{each layer $l = 0, \dots, L-1$}
		\FOR{each function $\phi = 1, \dots, n_l\cdot n_{l+1}$}
		\STATE $\mu^\prime\left(\left\{c_i^{(l,\phi)}\right\}_{i=1}^{G^\prime+k}\right) \gets \text{Interp}\left[\mu\left(\left\{c_i^{(l,\phi)}\right\}_{i=1}^{G+k}\right)\right]$
		\STATE $\nu^\prime\left(\left\{c_i^{(l,\phi)}\right\}_{i=1}^{G^\prime+k}\right) \gets \text{Interp}\left[\nu\left(\left\{c_i^{(l,\phi)}\right\}_{i=1}^{G+k}\right)\right]$
		\ENDFOR
		\ENDFOR
		
		\RETURN $\left(t^\prime, \mu^\prime, \nu^\prime\right)$
	\end{algorithmic}
\end{algorithm}

Instead of re-initializing the optimizer's state, the values for the moments are kept fixed for the parameters that are not affected by the extension of the grid, namely $c_r$ and $c_B$. As far as $\left\{c_i\right\}_{i=1}^{G+k}$ are concerned, their number increases due to the grid's extension, so their moments cannot be retained ``as-is'' in the new state. However, rather than resetting them, one may assign values to them by performing a linear interpolation based on the optimizer's state prior to the grid's extension. This process is analogous to using the least squares algorithm after grid extensions to get good estimates for the values of the $\left\{c_i\right\}_{i=1}^{G+k}$ parameters.

\begin{figure}[!t]
	\centering
	\includegraphics[width=\linewidth]{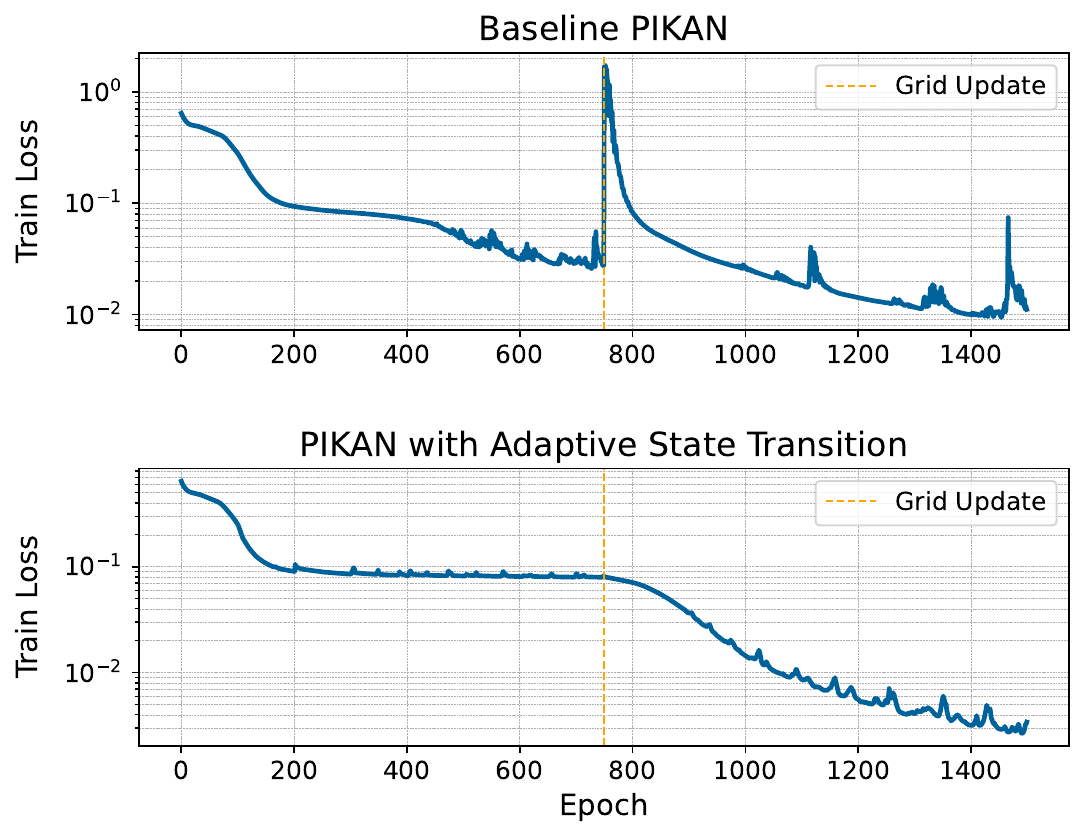}
	\caption{Training a PIKAN without (top) and with (bottom) the adaptive state transition technique.}
	\label{adapt-res}
\end{figure}

We implemented this algorithm in \texttt{jaxKAN} using \texttt{jax.numpy.interp} for the linear interpolation and \texttt{jax.vmap} to parallelize the interpolation across functions within each layer. During experimentation we found that incorporating Nesterov momentum into Adam \cite{nesterov} makes the state transition even smoother. To demonstrate how this technique affects the training of PIKANs after grid extensions, we trained an additional $[2, 8, 8, 1]$ PIKAN for Burgers' equation, using the same parameters. In this case, we performed regular grid adaptations with a period of 100 epochs prior to the grid's extension, decreased the learning rate by a factor of 0.3 following the grid's update and adopted Algorithm \ref{alg:state_transition} to update the optimizer's state after the extension. The results can be seen in the lower graph of Fig. \ref{adapt-res}. The loss function's values for the first PIKAN are also depicted in the upper graph of the same figure to serve as a baseline. Evidently, the application of the technique completely removes the sharp increase of the loss function's values. More importantly, the final value of the loss function is considerably lower compared to its value after 1500 epochs of training the first PIKAN. This indicates that the adaptive state transition not only solves the problem of the sharp peaks, but also enhances the network's training.

It should be noted that this adaptive state transition is neither exclusive to Adam nor exclusive to spline basis functions. In fact, it can even be applied to problems involving KANs outside the field of PDEs. In Appendix \ref{appB}, we present one such example, where it is used for KANs that are trained to learn functions. There, the technique's effect on the loss function is much more prominent than in the case of PIKANs, not merely smoothing the sharp increases in the loss function's value but turning them to sharp decreases instead.

\subsection{Loss re-weighting}

The next method for adaptive training is related to the loss function's terms and has already been applied for PINNs with MLPs, as well as PIKANs \cite{karniadakisreview}. This method, known as residual-based attention (RBA), was introduced in \cite{rba}. Within the context of RBA, additional terms are introduced in \eqref{eq3} and \eqref{eq4}, transforming them to:

\begin{align}
	\mathcal{L}_f\left(\theta\right) &= \frac{1}{N_f}\sum_{i=1}^{N_f}{\lVert \alpha_{0,i}\left( \mathcal{F}\left(u\left(x_i;\theta\right)\right) - f\left(x_i\right) \right)\rVert^2}, \label{eq12} \\
	\mathcal{L}_b^k\left(\theta\right) &= \frac{1}{N_b^k}\sum_{i=1}^{N_b^k}{\lVert \alpha_{k,i}\left( \mathcal{B}^k\left(u\left(x_i;\theta\right)\right) - b^k\left(x_i\right) \right)\rVert^2}, \label{eq13}
\end{align}

\noindent where $\left\{\alpha_{k,i}\right\}_{k=0}^{n_\Gamma}$ are self-adaptive weights that, unlike the ``global'' weights $w_f$ and $\left\{w_b^k\right\}_{k=1}^{n_\Gamma}$, define local scalings unique to each collocation point. Their initial value is $1.0$ and they are updated based on the rule:

\begin{equation}
	\alpha_{k,i}^{\left(t+1\right)} \gets \left(1-\eta\right)\alpha_{k,i}^{\left(t\right)} + \eta\frac{\left|r_k\left(x_i\right)\right|}{\max_i\left(\left\{\left|r_k\left(x_i\right)\right|\right\}\right)}, \label{eq14}
\end{equation}

\noindent where $x_i$ is the $i$-th collocation point's coordinate vector, $t$ is the training epoch's index, $0 \leq \eta \leq 1$ is a mixing factor and

\begin{align}
	r_k\left(x_i\right) = 
	\begin{cases}
		\mathcal{F}\left(u\left(x_i;\theta\right)\right) - f\left(x_i\right),~\text{ for }~ k = 0 \\
		\mathcal{B}^k\left(u\left(x_i;\theta\right)\right) - b^k\left(x_i\right), ~\text{ for }~ k > 0
	\end{cases} \label{eq15}
\end{align}

\noindent is the $i$-th collocation point's residual with respect to the PDE ($k = 0$) or the boundary conditions ($k>0$).

\begin{figure}[!t]
	\centering
	\includegraphics[width=\linewidth]{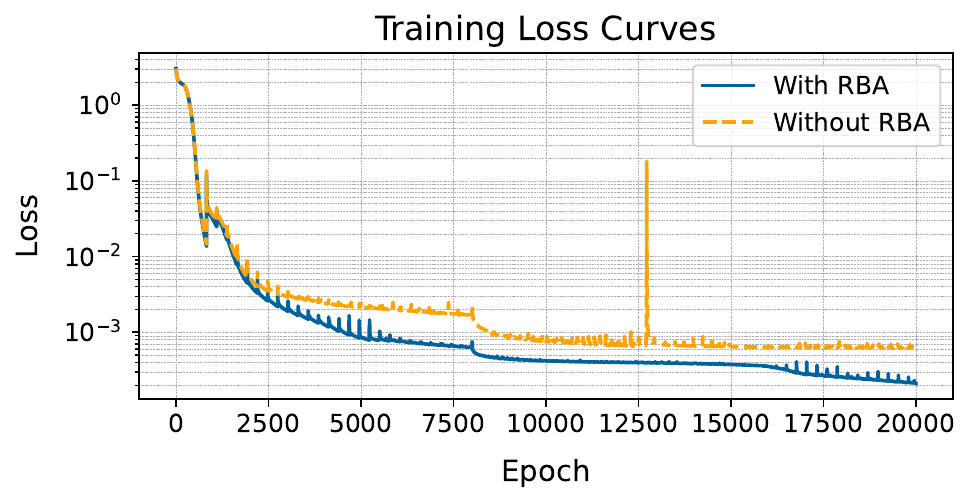}
	\caption{Training loss of a PIKAN for the Allen-Cahn equation with (blue, solid line) and without (orange, dashed line) RBA.}
	\label{rba}
\end{figure}

RBA is an adaptive loss re-weighting scheme where the contribution of each collocation point to the overall loss function is scaled based on the value of its residual. Fig. \ref{rba} demonstrates an example of applying RBA with $\eta = 10^{-4}$, which is the value used for all subsequent experiments presented herein, to train a $[2,8,8,1]$ PIKAN for the Allen-Cahn equation. An identical PIKAN without RBA is used as the baseline. During the first 2000 epochs, the two PIKANs appear identical in terms of their training. However, as the values of the $\alpha$-weights start adapting, it becomes evident that the PIKAN with RBA trains faster. It is noted that a grid extension $G = 3 \to G^\prime = 8$ occurs for both PIKANs at the 8000th training epoch, although it does not lead to a sharp loss increase since the technique presented in the previous subsection is used.

\subsection{Collocation points re-sampling}

\begin{figure}[!t]
	\centering
	\includegraphics[width=\linewidth]{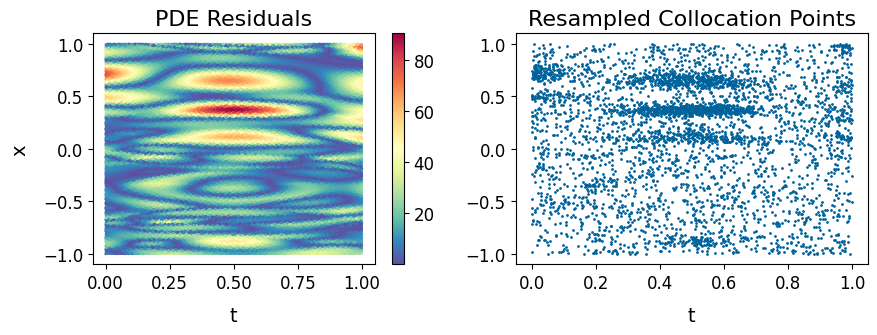}
	\caption{Example application of RAD during the training of a PIKAN for the Helmholtz equation.}
	\label{rad}
\end{figure}

The final adaptive training technique has also been used to train MLP-based PINNs, however its application for PIKANs requires caution. It involves re-sampling collocation points during training using a residual-based adaptive distribution (RAD) \cite{colloc2}, which may interact with other adaptive methods employed simultaneously, including the grid itself. The main idea of RAD is to dynamically change the set of collocation points by making their distribution denser in regions of the domain where the absolute values of the PDE residuals are high and sparser in regions where they are close to zero. To achieve this, we define:

\begin{equation}
	p\left(x_i\right) = \frac{\left|r_0^a\left(x_i\right)\right|}{\frac{1}{N_f}\sum_{i=1}^{N_f}{\left|r_0^a\left(x_i\right)\right|}} + c, \label{eq16}
\end{equation}

\noindent where $r_0\left(x_i\right)$ is the PDE residual given by \eqref{eq15} and $a \geq 0$, $c \geq 0$ are the RAD method's hyper-parameters. Based on \eqref{eq16}, the adaptive resampling is performed by sampling a set of dense points, $\mathcal{S}$, within the PDE's domain (much denser than the set of collocation points) and then computing $p\left(x_i\right)$ for every $x_i \in \mathcal{S}$. The new set of collocation points corresponds to $N_f$ points sampled from $\mathcal{S}$, according to $p\left(x\right)$. The implementation of this procedure for \texttt{jaxKAN} is described by Algorithm \ref{alg:collocs} and an example application for $a = 3$ and $c = 1$ can be seen in Fig. \ref{rad}, from a snapshot taken from the 1000th training epoch of a $[2,8,8,1]$ PIKAN trained to solve the Helmholtz equation.

\begin{algorithm}[H]
	\caption{Adaptive Resampling of Collocation Points}
	\label{alg:collocs}
	\begin{algorithmic}[1]
		\renewcommand{\algorithmicrequire}{\textbf{Input:}}
		\renewcommand{\algorithmicensure}{\textbf{Output:}}
		
		\REQUIRE $N_f$, parameters $a, c \geq 0$, weights $\left\{\alpha_{0,i}\right\}$, grid size $G$
		\ENSURE Set of new collocation points $\mathcal{T}$, new RBA weights $\left\{\alpha^\prime_{0,i}\right\}$, new grid $\mathcal{G}^\prime$
		
		\STATE Sample a set of dense points $\mathcal{S} \subseteq \Omega$ with $\left|\mathcal{S}\right| \gg N_f$
		\FOR{each point $x_i \in \mathcal{S}$}
		\STATE Calculate $p\left(x_i\right)$ using \eqref{eq15} and \eqref{eq16}
		\ENDFOR
		
		\STATE $p\left(x\right) \gets p\left(x\right) / \sum_{x_i \in \mathcal{S}}{p\left(x_i\right)}$
		
		\STATE Sample $N_f$ points from $\mathcal{S}$ according to $p\left(x\right)$:
		\STATE \quad $\left\{i\right\}_{i=1}^{N_f} \gets$ \texttt{jax.random.choice} based on $p\left(x\right)$
		\STATE \quad $\mathcal{T} \gets \mathcal{S}\left[\left\{i\right\}_{i=1}^{N_f}\right]$
		
		\STATE Re-initialize RBA weights:
		\STATE \quad $\bar{\alpha} \gets \frac{1}{N_f}\sum_{i=1}^{N_f}{\alpha_{0,i}}$
		\FOR{$i = 1, \dots, N_f$}
		\STATE $\alpha^\prime_{0,i} \gets \bar{\alpha}$
		\ENDFOR
		
		\STATE Perform grid adaptation using $\mathcal{T}, G$:
		\STATE \quad $\mathcal{G}^\prime \gets$ Equation \eqref{eq11}
		
		\RETURN $\mathcal{T}$, $\left\{\alpha^\prime_{0,i}\right\}$, $\mathcal{G}^\prime$
	\end{algorithmic}
\end{algorithm}

Clearly, the introduction of the RAD method directly impacts the RBA technique, which is why the two methods have not been used conjointly for the training of PINNs. Specifically for PIKANs, RAD also has a direct impact to the grid, as adaptive grids depend on the distribution of the collocation points, and therefore the basis functions themselves. In order to incorporate both methods such that they do not adversely impact each other, careful fine-tuning is necessary. Since each RBA weight corresponds to a distinct collocation point, resampling the collocation points necessitates re-initializing the RBA weights so their training can begin anew using \eqref{eq14}. Consequently, choosing the optimal epochs to perform re-sampling is crucial; frequent applications of RAD may render RBA ineffective, while infrequent applications may not be sufficient for the method to work effectively. As far as the re-initialization of the RBA weights is concerned, we found that starting from the mean value of the set's trained weights works well, although more sophisticated transition methods can be devised. Regarding the grid, an adaptation is required after applying RAD to ensure it is updated based on the new set of collocation points.

\subsection{Results with adaptive training}

\begin{figure*}[!t]
	\centering
	\includegraphics[width=\linewidth]{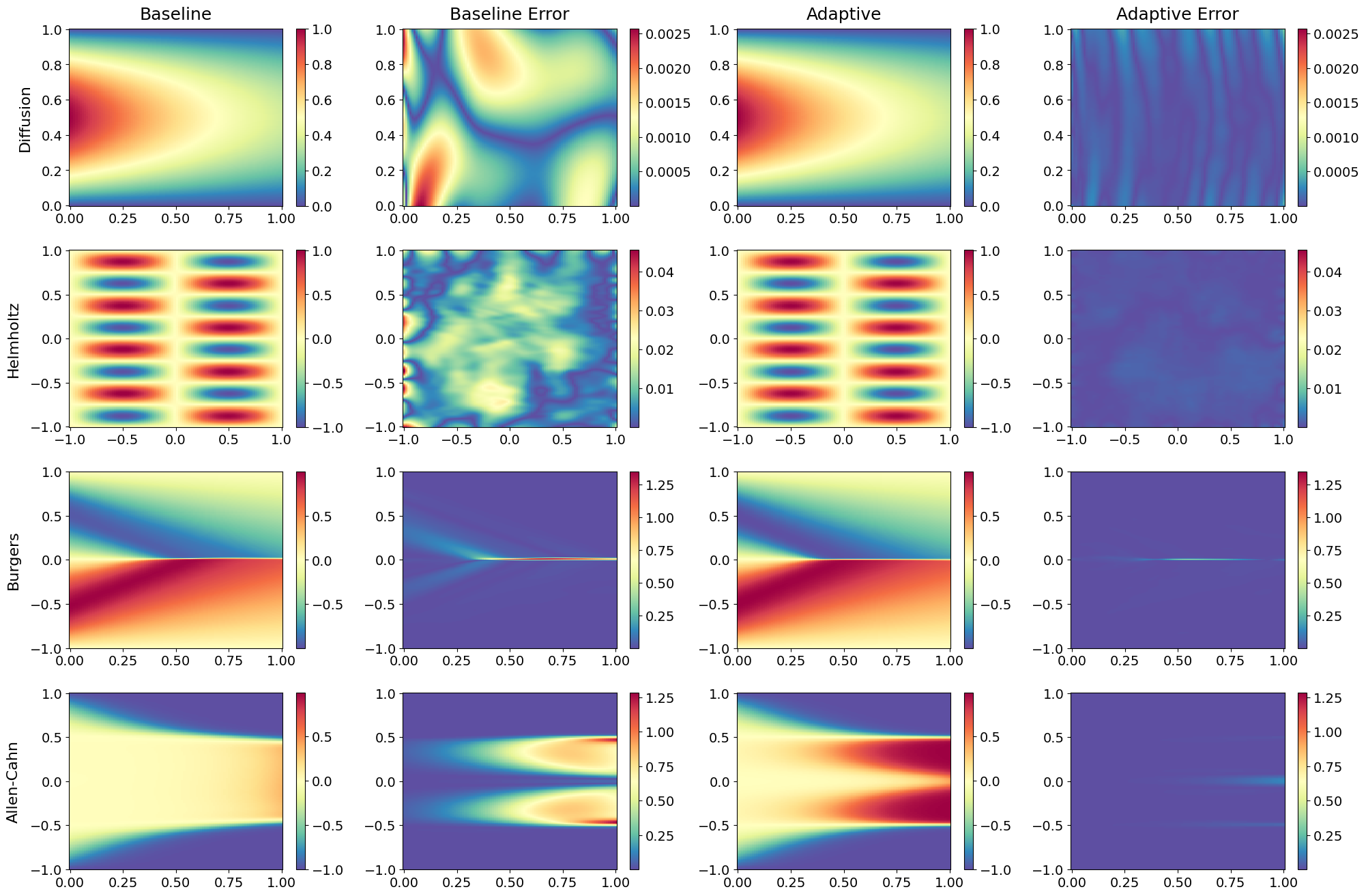}
	\caption{PIKAN results for the Diffusion equation (first row), the Helmholtz equation (second row), Burgers' equation (third row), and the Allen–Cahn equation (fourth row). The first/third and second/fourth columns correspond to the baseline/adaptive result and its absolute error compared to the reference solution, respectively. In each row, the solutions/errors share the same colorbars.}
	\label{adaptive-pikan}
\end{figure*}

\begin{table*}[!t]
	\caption{\textbf{Comparison of PIKAN training parameters and performance metrics with \cite{karniadakisreview} for the Helmholtz and Allen-Cahn equations.}}
	\centering
	\setlength{\tabcolsep}{12pt}
	%\begin{tabular}{|p{0.2\linewidth}|p{0.2\linewidth}|p{0.2\linewidth}|}
	\begin{tabular}{|c|c|c|c|c|c|}
		\hline
		\rule{0pt}{3ex} PDE & Implementation & Parameters Number & Epochs & Relative L$^2$ Error & Time (ms/epoch) \\ [0.5ex]
		\hline
		\rule{0pt}{3ex} Helmholtz & \texttt{jaxKAN} & 1848 & $10^5$ & 0.176\% & 18.3 \\
		\rule{0pt}{3ex}  &  \cite{karniadakisreview} & 15840 & $2\cdot10^5$ & 0.16\% & 7.4 \\ [0.5ex]
		\hline
		\rule{0pt}{3ex} Allen-Cahn & \texttt{jaxKAN} & 1584 & $10^5$ & 1.414\% & 12.1 \\
		\rule{0pt}{3ex}  & \cite{karniadakisreview} & 6720 & $1.5\cdot10^5$ & 5.15\% & 39.31 \\ [0.5ex]
		\hline
	\end{tabular}
	\label{comparisons}
\end{table*}

Using all the aforementioned adaptive methods or combinations thereof, we revisited the four PDEs studied in Section \ref{sec:framework} to acquire new benchmarks. First, we trained $[2,8,8,1]$ PIKANs without adaptive features or grid updates for $10^5$ epochs with a learning rate of $10^{-3}$ to establish baselines for each PDE. For their grids, we chose an initial value of $G = 3$. Additionally, we opted for $g_e = 0.05$, ensuring grids that are practically fully adaptive. Next, we trained PIKANs adaptively for the same number of epochs, using different configurations per PDE which are detailed in the following. The results are presented in Fig. \ref{adaptive-pikan}.

\begin{itemize}
	\item \textbf{Diffusion Equation}: Starting from a learning rate of $10^{-3}$, it was progressively scaled down to $8.75\cdot 10^{-3}$ following grid extensions. Grid adaptations were performed every 275 epochs and two grid extensions were performed for $G^\prime = 8$ and $G^\prime = 14$. RBA was used throughout training, but RAD was not utilized. The relative L$^2$ error for the baseline is equal to 0.193\%, while the relative L$^2$ error for the adaptive PIKAN is equal to 0.021\%. \\
	
	\item \textbf{Helmholtz Equation}: Starting from a learning rate of $10^{-3}$, it was progressively scaled down to $2.5\cdot 10^{-3}$ following grid extensions. Grid adaptations were performed every 200 epochs and two grid extensions were performed for $G^\prime = 7$ and $G^\prime = 15$. RBA was used throughout training, but RAD was not utilized. For both the baseline and the adaptive PIKAN, a global weight $w_f = 0.01$ was used, following \cite{karniadakisreview}. The relative L$^2$ error for the baseline is equal to 2.069\%, while the relative L$^2$ error for the adaptive PIKAN is equal to 0.176\%. \\
	
	\item \textbf{Burgers' Equation}: Starting from a learning rate of $10^{-3}$, it was progressively scaled down to $1.3\cdot 10^{-4}$ following grid extensions. Grid adaptations were performed every 300 epochs and two grid extensions were performed for $G^\prime = 8$ and $G^\prime = 14$. RBA was used throughout training, and RAD was performed twice with $a = c = 1$. The relative L$^2$ error for the baseline is equal to 13.504\%, while the relative L$^2$ error for the adaptive PIKAN is equal to 2.430\%. \\
	
	\item \textbf{Allen-Cahn Equation}: Starting from a learning rate of $10^{-3}$, it was progressively scaled down to $1.5\cdot 10^{-4}$ following grid extensions. Grid adaptations were performed every 275 epochs and two grid extensions were performed for $G^\prime = 8$ and $G^\prime = 12$. RBA was used throughout training, and RAD was performed three times with $a = c = 1$. The relative L$^2$ error for the baseline is equal to 44.430\%, while the relative L$^2$ error for the adaptive PIKAN is equal to 1.414\%. \\
\end{itemize}

It is noted that the aforementioned results are not necessarily the best-obtainable using the presented adaptive training techniques; careful fine-tuning and extensive testing may lead to even better benchmarks. Nonetheless, the relative L$^2$ errors achieved in our work are already comparable to or better than these reported in \cite{karniadakisreview}, albeit with PIKANs that are trained for a shorter number of epochs and with considerably fewer parameters. Table \ref{comparisons} compares our results to the ones obtained in \cite{karniadakisreview} for the Helmholtz and Allen-Cahn equations, which can be considered the state-of-the-art for PIKANs, due to the currently limited literature on the subject. The diffusion equation was not studied in \cite{karniadakisreview}. With regard to Burgers' equation, although \cite{karniadakisreview} report a minimum relative L$^2$ error of 2.71\%, which is higher than our result of 2.43\%, a direct comparison would be unfair, due to the authors' using a different method to solve this equation. Importantly, even though a better GPU was used in \cite{karniadakisreview} and more collocation points were utilized, our approach is comparable even in terms of training times. This is unprecedented for PIKANs with grid extension during training, especially taking into account that the latencies incurred by the compilation of some functions are included in our reported times.

\section{Grid-Dependent Basis Functions} \label{sec:grid}

The experiments presented thus far demonstrate how promising KANs are as substitutes for MLPs in PINNs, particularly when trained in an adaptive manner. Nevertheless, their training time remains a significant disadvantage compared to MLPs. To address this, recent studies have focused on improving KAN efficiency by replacing the computationally expensive calculation of B-splines with other basis functions. While this is theoretically a sound solution, a poor choice of alternative basis functions may indeed accelerate the training time of KANs (and PIKANs by extension), but at a severe cost to their performance. In the following, we focus on the distinction of candidate replacements for spline basis functions based on their dependency on a grid. In particular, we introduce the concepts of staticity and full grid adaptivity for basis functions and demonstrate the advantages of opting for basis functions that are both non-static and fully adaptive to grids when training PIKANs.

\subsection{Staticity}

Jacobi polynomials, which have been extensively used as basis functions for KANs, including for the training of efficient and accurate PIKANs \cite{karniadakisreview,fkan}, fall under the category of grid-independent basis functions. They are univariate functions which can be constructed via a Gaussian hypergeometric function or equivalent recurrence relations. Their inherent independence from a grid classifies Jacobi-based KANs as ``static'', because it prevents them from exploiting the dynamic grid extension method, which has been demonstrated to enhance efficiency by progressively fine-graining their basis functions during the training process. Notably, in Section \ref{sec:training}, we showed that this effect can be amplified via the adaptive optimizer state transition technique, where changing the grid's size leads to an immediate, sharp reduction in the loss function's values, even if they had reached a plateau prior to the grid's extension.

In contrast, KANs with non-static (i.e., grid-dependent) basis functions do not have such restrictions. In FastKAN \cite{rbfkan}, which is an implementation of KANs utilizing radial basis functions, the dependency on a grid arises from the choice of centers for the set of radial basis functions. ReLU-KANs \cite{relukan} employ $R_i\left(x\right)$ as basis functions, with

\begin{equation}
	R_i\left(x\right) = \left[r_i\cdot\text{ReLU}\left(e_i-x\right)\cdot\text{ReLU}\left(x-s_i\right)\right]^2, \label{eq17}
\end{equation}

\noindent where $r_i = 4\left(e_i-s_i\right)^{-2}$ is a normalization constant and $\text{ReLU}\left(x\right) = \max\left(0,x\right)$. Evidently, their dependence on $e_i$ and $s_i$ constitutes $R_i\left(x\right)$ as non-static basis functions. Finally, in Wav-KANs \cite{wavkan}, the grid dependency varies with different types of wavelets. For instance, if one defines a basis of Ricker (also known as ``Mexican hat'') wavelets 

\begin{equation}
	\psi_i\left(x\right) = 2\cdot\frac{1-\left(\frac{x-\mu_i}{\sigma_i}\right)^2}{\sqrt{3\sigma_i}\pi^{1/4}}\exp\left(-\frac{\left(x-\mu_i\right)^2}{2\sigma_i^2}\right), \label{eq18}
\end{equation}

\noindent then the grid dependency comes from the $\mu_i$, $\sigma_i$ parameters. It is noted that, even though the studies that introduced them employed static grids in all of their experiments, this staticity resulted from choosing not to perform grid updates rather than an inherent limitation of the basis functions. %Nonetheless, instead of defining a basis, a single wavelet was employed to replace the B-spline curve in the paper presenting Wav-KANs. This is equivalent to using a grid with a single interval ($G=1$).

\subsection{Full Grid Adaptivity}

\begin{figure}[!t]
	\centering
	\includegraphics[width=\linewidth]{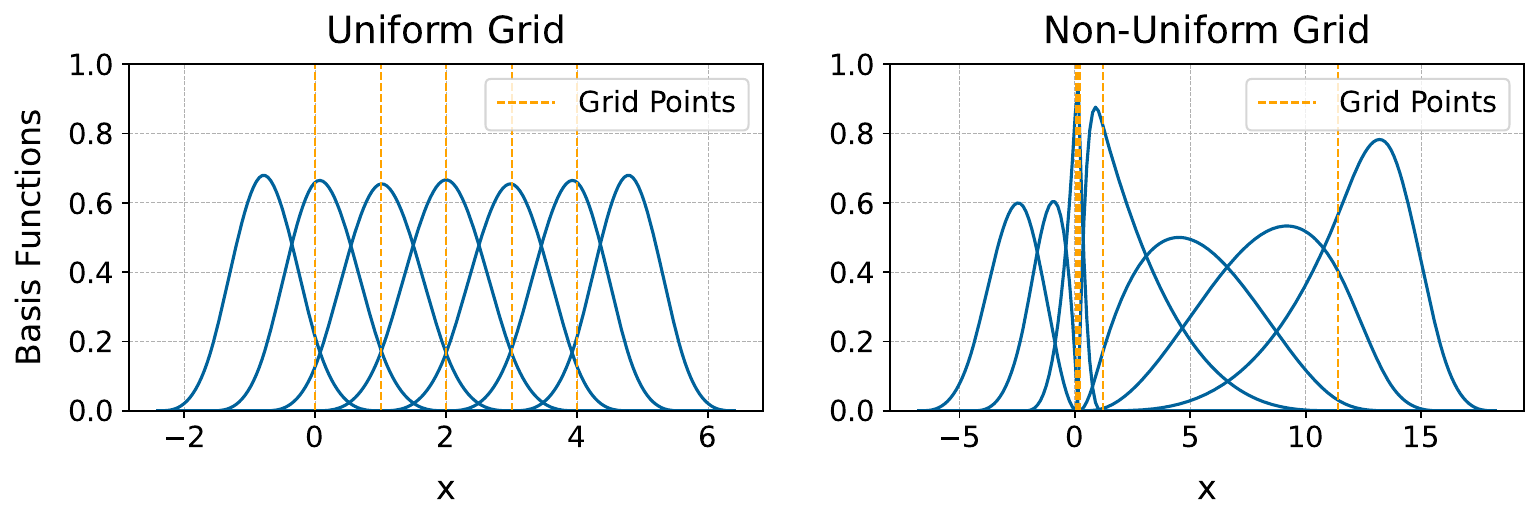}
	\caption{Example of spline basis functions of order $k$ = 3 calculated on a uniform (left) and a non-uniform (right) grid.}
	\label{spline-basis}
\end{figure}

In addition to being non-static, it is important for candidate basis functions to also be fully adaptive to the grid in order to constitute viable alternatives to B-Splines. Full grid adaptivity means that uniform grids should give rise to uniform basis functions, while non-uniform grids should produce basis functions that are denser in more dense regions and sparser in less dense regions of the grid. Simultaneously, it is essential to ensure that there are no regions within the grid where all basis functions are equal to zero. Fully adaptive basis functions are essential because, as discussed in the context of adaptive sampling of collocation points, the grids in KANs can be designed to adapt to the data and their distribution. Therefore, fully adaptive basis functions are, by extension, adaptive to the problem's data themselves.

The concept of full grid adaptivity is demonstrated in Fig. \ref{spline-basis} for spline basis functions of order $3$, where $\mathcal{G}_1 = \left[0.0, 1.0, 2.0, 3.0, 4.0\right]$ is used as an example of a uniform grid and $\mathcal{G}_2 = \left[0.05, 0.14, 0.22, 1.20, 11.40\right]$ is used as an example of a non-uniform grid. The left graph of Fig. \ref{spline-basis} depicts the uniform basis functions constructed on the uniform grid. Conversely, the basis functions for the non-uniform grid shown in the right graph of Fig. \ref{spline-basis} are denser in the region around $0.0$, reflecting the higher concentration of grid points in this area. Notably, there is no region inside the grid where all spline basis functions are simultaneously zero. Note that, even though $G=4$ in both cases, the total number of spline basis functions is $7$ due to the grid augmentation that occurs prior to their construction \cite{pykan}. During this augmentation process, additional support points are appended at the ends of the grid, so that a grid with $G$ intervals corresponds to $G+k$ spline basis functions of order $k$.

\subsection{The case study of ReLU-KANs}

Unlike spline basis functions, for which we needed to artificially tweak training parameters in order to simulate staticity and non-full grid adaptivity, the $\left\{R_i\right\}$ basis introduced for ReLU-KANs is a clear example of non-fully adaptive basis functions by design. In \cite{relukan}, the $R$ basis functions were constructed to be equidistant on a uniform grid by choosing $s_i = \left(i-k-1\right)/G$ and $e_i = i/G$, therefore no experiments were performed on non-uniform grids. However, there are numerous PDE problems where the corresponding grids are not uniform, such as when adding more collocation points in a specific region to impose a boundary condition. In such cases, a grid with many intervals (large $G$) would be required for the uniform $R$ basis functions to provide the proper support in denser regions of the grid, thus making the switch to $R$ basis functions for higher efficiency counterproductive. For this reason, we propose a different construction method to make the $R$ basis functions fully adaptive to a generally non-uniform grid, described by Algorithm \ref{alg:rbasis}.

\begin{algorithm}[H]
	\caption{Grid-Adaptive $R$ Basis Functions Construction}
	\label{alg:rbasis}
	\begin{algorithmic}[1]
		\renewcommand{\algorithmicrequire}{\textbf{Input:}}
		\renewcommand{\algorithmicensure}{\textbf{Output:}}
		
		\REQUIRE Grid $\mathcal{G} \in \mathbb{R}^{G+1}$, parameters $p, k \in \mathbb{N}$
		\ENSURE $R$ basis function parameters $\left\{s_i, e_i, r_i\right\}_{i=1}^{G+1}$
		
		%\STATE \textbf{Augment Grid}
		\STATE Initialize $\tilde{\mathcal{G}} \gets \mathcal{G}$
		\FOR{$i = 1, \dots, p$}
		\STATE Append a point at each end of $\tilde{\mathcal{G}}$:
		\STATE $p_s \gets \tilde{\mathcal{G}}[0] - \frac{1}{k}\left(\tilde{\mathcal{G}}[k] - \tilde{\mathcal{G}}[0]\right)$
		\STATE $p_e \gets \tilde{\mathcal{G}}[G] + \frac{1}{k}\left(\tilde{\mathcal{G}}[G] - \tilde{\mathcal{G}}[G-k]\right)$
		\STATE $\tilde{\mathcal{G}} \leftarrow \left\{p_s\right\} \cup \tilde{\mathcal{G}} \cup \left\{p_e\right\}$
		\ENDFOR
		
		%\STATE \textbf{Calculate $(e, s)$ Values}
		\FOR{$i = p, \dots, G + p$}
		\STATE $s_i \gets \tilde{\mathcal{G}}[i] - 0.5\cdot\left(\tilde{\mathcal{G}}[i+p] - \tilde{\mathcal{G}}[i-p]\right)$
		\STATE $e_i \gets 2\cdot \tilde{\mathcal{G}}[i] - s_i$
		\STATE $r_i \gets 4\cdot \left(e_i - s_i\right)^{-2}$
		\ENDFOR
		
		\RETURN $\left\{s_i, e_i, r_i\right\}_{i=1}^{G+1}$
	\end{algorithmic}
\end{algorithm}

\begin{figure}[!t]
	\centering
	\includegraphics[width=\linewidth]{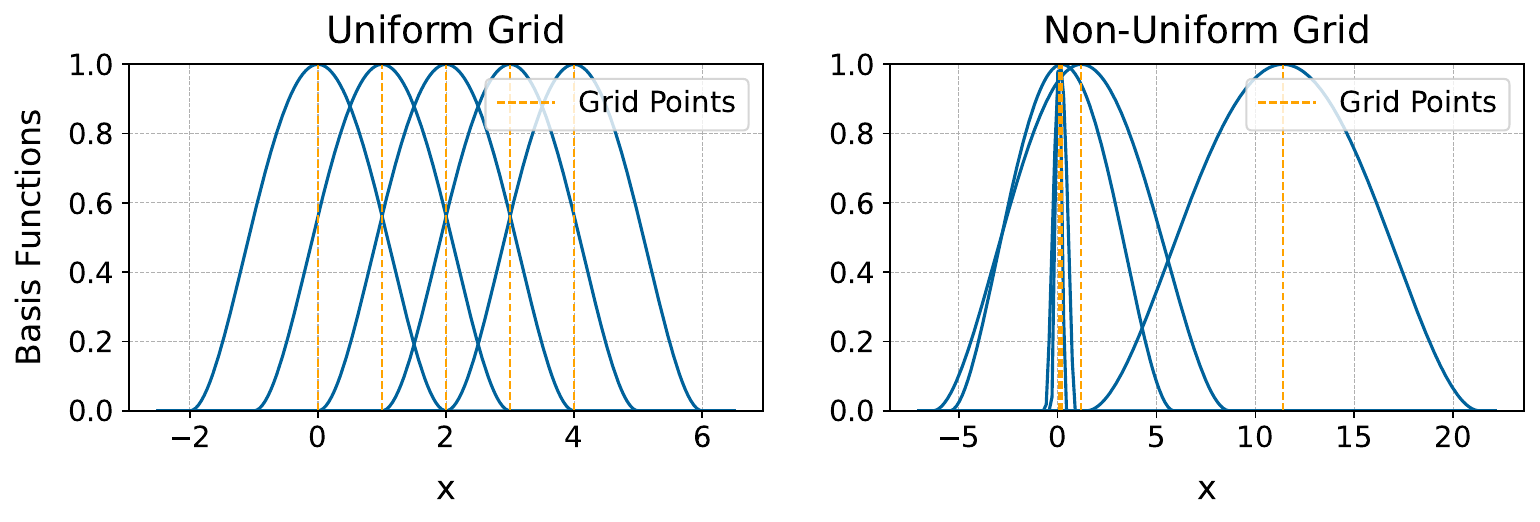}
	\caption{Example of $R$ basis functions calculated on a uniform (left) and a non-uniform (right) grid for $p = 2$ and $k = 3$.}
	\label{r-basis}
\end{figure}

\begin{figure*}[!t]
	\centering
	\includegraphics[width=0.8\linewidth]{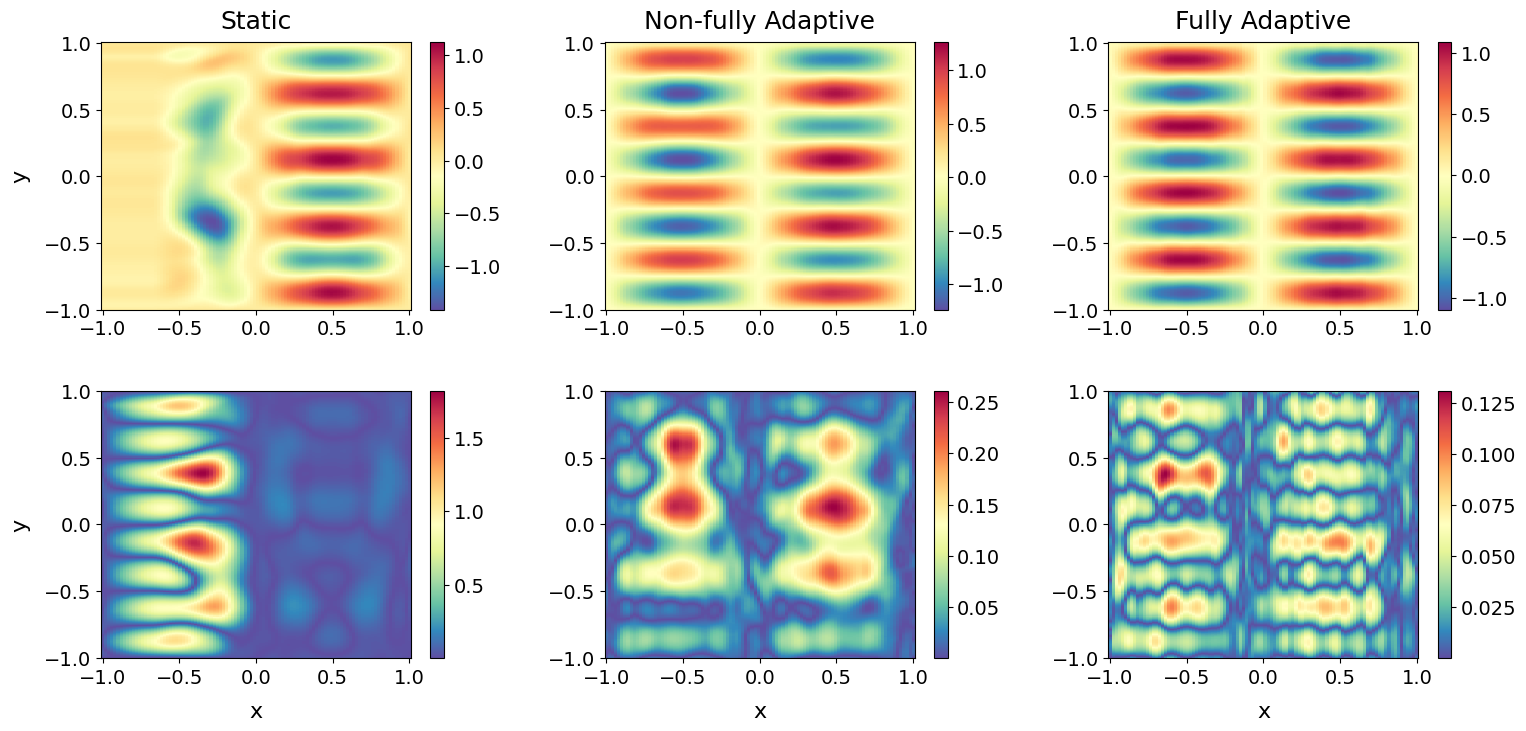}
	\caption{PIKAN results with $R$ basis functions for the Helmholtz equation with staticity (first column), non-full adaptivity (second column) and full adaptivity (third column). The upper row depicts each experiment's result, while the lower row shows each experiment's absolute error compared to the reference solution.}
	\label{relukan-results}
\end{figure*}

First, we center each $R_i\left(x\right)$ around a grid point, $\mathcal{G}[i]$, by imposing $s_i + e_i = 2\mathcal{G}[i]$. This ensures that all grid points have support from the basis functions, the density of which follows the density of the grid. Next, to ensure that there are no regions within the grid with zero support, we set $s_i=\mathcal{G}[i]-0.5\left(\mathcal{G}[i+p]-\mathcal{G}[i-p]\right)$, where $p$ is a parameter controlling how many neighboring grid points are taken into account when defining the coverage (width) of the basis functions. To apply this condition for points near the grid's edges, we perform an augmentation technique analogous to the one used for spline basis functions. Specifically, we extend the grid by $p$ points at both ends, recursively taking into account the distance between each edge point and its $k$-th neighbor, normalized by $k$. This construction ensures that $G+1$ basis functions are obtained for a grid with $G$ intervals. The $R$ basis functions constructed using this process with $p=2$ and $k=3$ for $\mathcal{G}_1$ and $\mathcal{G}_2$ can be seen in the left and right graphs of Fig. \ref{r-basis}, respectively.

It should be noted that this construction redefines the parameter $k$, which had no physical meaning in the original implementation of ReLU-KANs other than providing a way to directly compare $R$ basis functions to spline basis functions on a uniform grid, both qualitatively and quantitatively. Importantly, this construction is by no means the only method to make $R$ basis functions fully adaptive. Nevertheless, it is intuitive and can serve as the basis for a broader methodology of constructing non-static, fully adaptive basis functions. For instance, extending these ideas to radial basis functions or Ricker wavelets is straightforward: one may choose centers for the former and $\mu_i$ for the latter which lie on the grid points' coordinates and then adjust their spread to ensure full coverage within the grid.

We implemented ReLU-KANs in \texttt{jaxKAN} based on this construction, utilizing just-in-time compilation and vectorized operations for improved computational efficiency. To confirm the importance of non-staticity and full grid adaptivity for the case of ReLU-KANs, we followed an approach similar to the one presented in the previous subsection. Specifically, we trained three $[2,8,8,1]$ PIKANs for $10^5$ epochs using the $R$ basis functions with $k = p = 2$ to solve the Helmholtz equation. For the first PIKAN, we imposed staticity by setting a constant $G = 3$ and $g_e = 1.0$. For the second PIKAN, we performed grid updates ($G = 3 \to 6 \to 12$) but only utilized uniform grids to impose non-full grid adaptivity ($g_e = 1.0$). For the third PIKAN, we performed grid updates ($G = 3 \to 6 \to 12$) and used grids adaptive to the data ($g_e = 0.05$), in order to investigate their impact in our fully adaptive implementation of $R$ basis functions. To ensure a fair comparison, all other parameters of the PIKANs were kept constant across all three experiments, with the exception of the learning rate: in the first case a constant learning rate was used, since no extension was performed, but for the other two cases the learning rate was scaled following grid extensions.

The solutions for each PIKAN along with their absolute errors from the reference solution are depicted in Fig. \ref{relukan-results}. The corresponding relative L$^2$ errors are 24.191\% for the static case, 6.761\% for the non-fully adaptive one and 5.552\% for the fully adaptive PIKAN, proving that non staticity and full grid adaptivity of their basis functions are important when training PIKANs. Notably, none of these PIKANs seem to outperform even the baseline set in Section \ref{sec:training} using spline basis functions. However, the average training time per epoch was (1.4 $\pm$ 0.1) ms and 3.4 ms for ReLU-KAN-based PIKANs without and with grid extensions, respectively. This indicates a significant tradeoff between performance and computational efficiency, although there are cases where this might be favorable. Ultimately, the choice of basis functions depends on whether the user favors speed or accuracy in their application. For instance, the adaptive PIKAN trained to solve Burgers' equation achieved the highest relative L$^2$ error, however this was solely due to the fact that the $x=0$ boundary was not identified with full precision. Nonetheless, the qualitative result of the existence of the discontinuity at $x=0$ was clear even through the baseline experiments.

\section{Summary \& Future Work} \label{sec:conclusion}

In the present study, we attempted to set the foundations for an adaptive training scheme for PIKANs with the purpose of improving their accuracy and optimizing their performance. To this end, we introduced an adaptive technique to smoothen the transition process after grid extensions, based on regular grid adaptations and a reconfiguration of the optimizer's state, without completely resetting it. Additionally, we adopted the RBA technique for adaptively re-weighting the loss function's terms, as well as the RDA technique to adaptively re-sample collocation points during training, both based on the residuals of the loss function. For the latter, we proposed an implementation that does not hinder the progress of RBA and conforms to the grid' adaptivity.

All of these methods were implemented in JAX within our newly introduced \texttt{jaxKAN} framework. Although our study focused on PIKANs, the framework itself can be utilized for a series of other applications. Practically any model that utilizes KANs as its underlying architecture can benefit from \texttt{jaxKAN}, with real-world applications spanning from molecular physics \cite{molecularkan} and quantum circuits \cite{quantumkan} to medical applications \cite{medicalkan} and cryptocurrencies \cite{cryptokan}. Regarding the PIKANs trained within our framework, they achieved training times which were 2 orders of magnitude faster than the PIKANs trained using \texttt{pykan}. Moreover, the performance of PIKANs which we trained adaptively to solve four different PDEs was on par with or superior to other types of PIKANs which were trained for larger numbers of epochs, with deeper and/or wider architectures. Importantly, the average training times were also comparable, thus constituting PIKANs as viable alternatives to MLP-based PINNs.

Finally, we proposed a general methodology for choosing alternative basis functions for PIKANs in cases when splines are too computationally inefficient. By defining the concept of staticity and full grid adaptivity, we advocated for utilizing basis functions that are able to cover the entire grid and provide support in specific regions based on the distribution of their inputs. Using ReLU-KAN-based PIKANs as a case study, we demonstrated that fully adaptive basis functions outperform static ones by comparing their accuracies for the solution of the Helmholtz PDE.

As far as future research in PIKANs is concerned, the field is ripe with numerous opportunities and new avenues to explore. Regarding the newly introduced \texttt{jaxKAN} framework, we intend to enrich it with additional KAN models and their corresponding basis functions, based on the attributes highlighted in Section \ref{sec:grid}. Moreover, studies to make the basis functions' grid dependency as efficient as possible are necessary. Ensuring that the basis functions cover the entire grid, with more support in denser areas, is a good starting point; however, further research is required, for instance, regarding their amplitudes or their desired overlap in dense/sparse regions. Another direction for future work is utilizing optimizers other than Adam with the adaptive state transition technique between grid updates. Finally, the overall adaptive training of PIKANs requires additional benchmarks, ideally followed by a strict mathematical formulation to understand how techniques like adaptive sampling of collocation points affect KANs differently from MLPs due to their inherent grid dependency. Generally, we believe that the end goal is to establish a set of mathematically-backed best practices for training PIKANs, ultimately creating a robust and unified pipeline for solving any type of differential equation.

\appendices
\section{Studied Partial Differential Equations} \label{appA}

\begin{figure}[!h]
	\centering
	\includegraphics[width=\linewidth]{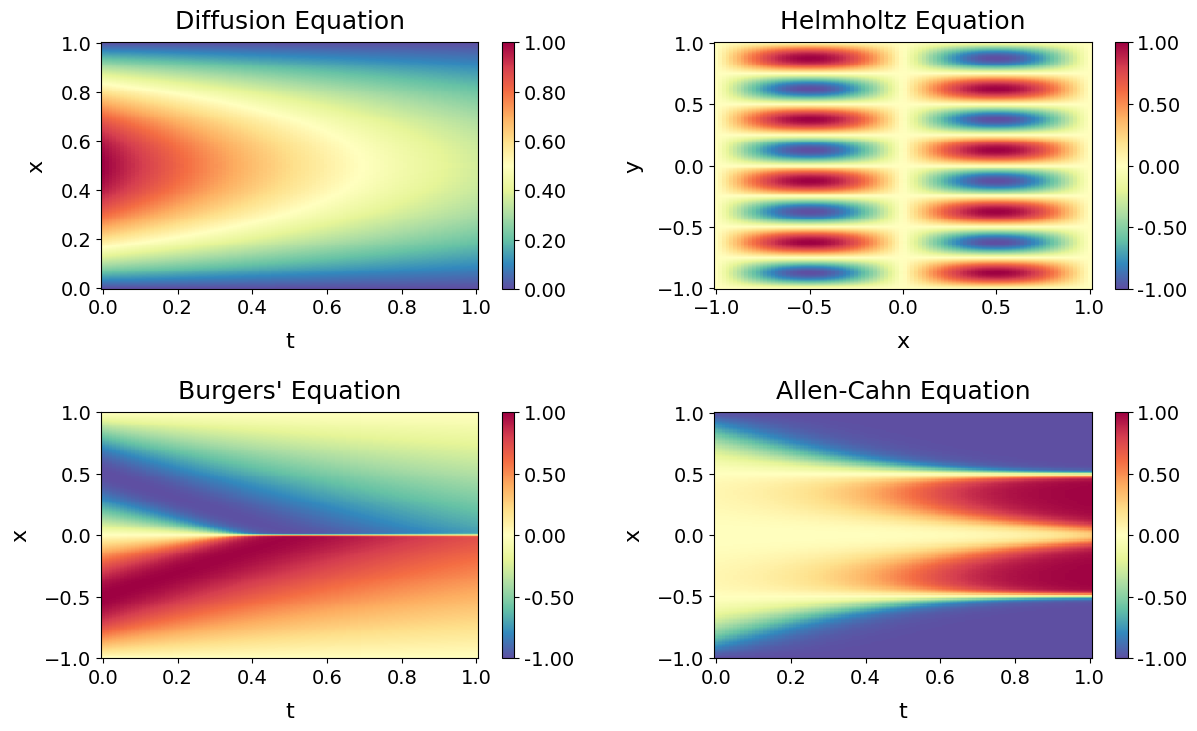}
	\caption{Reference solutions for each of the studied PDEs.}
	\label{pdes}
\end{figure}

The purpose of this Appendix is to present the four PDEs used in the experiments of this paper: the diffusion equation, the Helmholtz equation, Burgers' equation and the Allen-Cahn equation. Before presenting the PDEs with their boundary conditions, we note that for all PIKANs trained in the experiments presented in our work, a total of $N_f = 2^{12}$ collocation points were used for the PDE and $N_b = 2^6$ collocation points were used to impose each boundary condition. The reason why we opted for powers of two is because we sampled the collocation points from the Sobol sequence, a process mentioned in \cite{colloc2}. Additionally, the performance of the trained models was evaluated in terms of the relative L$^2$ error, defined as

% \cite{sobol}

\begin{equation*}
	L^2_\text{rel} = \frac{\lVert u_r\left(x\right) - u\left(x;\theta\right) \rVert}{\lVert u_r\left(x\right) \rVert},
\end{equation*}

\noindent where $u\left(x;\theta\right)$ is the PIKAN, $u_r\left(x\right)$ is the reference solution and $\lVert \cdot \rVert$ denotes the L$^2$ norm. The reference solution for each PDE is depicted in Fig. \ref{pdes}.

\subsection{Diffusion equation}

The 1+1-dimensional diffusion equation is:

\begin{equation*}
	\frac{\partial u}{\partial t} - \frac{\partial^2u}{\partial x^2} = f\left(t,x\right),
\end{equation*}

\noindent where $f\left(t,x\right)$ is the source term which was chosen equal to

\begin{equation*}
	f\left(t,x\right) = \left(\pi^2 - 1\right)\exp\left(-t\right)\sin\left(\pi x\right)
\end{equation*}

\noindent for the purposes of this paper. Additionally, $\Omega = \left[0,1\right]\times\left[0,1\right]$ is the chosen domain, with the following boundary conditions:

\begin{align}
	u\left(t=0,x\right) &= \sin\left(\pi x\right), \nonumber \\
	u\left(t,x=0\right) &= u\left(t,x=1\right) = 0 .\nonumber
\end{align}

\noindent The exact solution of this diffusion equation is

\begin{equation*}
	u\left(t,x\right) = \sin\left(\pi x\right)\exp\left(-t\right),
\end{equation*}

\noindent and is depicted in the upper left plot of Fig. \ref{pdes}.

\subsection{Helmholtz equation}

The 2-dimensional Helmholtz equation is defined as:

\begin{equation*}
	\frac{\partial^2 u}{\partial x^2} + \frac{\partial^2u}{\partial y^2} + k^2 u = f\left(x,y\right).
\end{equation*}

\noindent For our experiments we chose $k=1$ and

\begin{equation*}
	f\left(x,y\right) = \left(1 - 17\pi^2\right)\sin\left(\pi x\right)\sin\left(4\pi y\right),
\end{equation*}

\noindent with $\Omega = \left[-1,1\right]\times\left[-1,1\right]$ and boundary conditions

\begin{align}
	u\left(x=-1,y\right) &= u\left(x=1,y\right) = 0, \nonumber \\
	u\left(x,y=-1\right) &= u\left(x,y=1\right) = 0. \nonumber
\end{align}

\noindent This PDE also has an exact solution, which is

\begin{equation*}
	u\left(t,x\right) = \sin\left(\pi x\right)\sin\left(4 \pi y\right)
\end{equation*}

\noindent and is depicted in the upper right plot of Fig. \ref{pdes}.

\subsection{Burgers' equation}

As far as Burgers' equation is concerned, it is defined as:

\begin{equation*}
	\frac{\partial u}{\partial t} + u\frac{\partial u}{\partial x} - \nu \frac{\partial^2u}{\partial x^2} = 0
\end{equation*}

\noindent in the $\Omega = \left[-1,1\right]\times\left[0,1\right]$ domain. For the boundary conditions

\begin{align}
	u\left(t=0,x\right) &= -\sin\left(\pi x\right), \nonumber \\
	u\left(t,x=-1\right) &= u\left(t,x=1\right) = 0 ,\nonumber
\end{align}

\noindent the solution of Burgers' equation does not have an analytical expression. For this reason, we used the reference solution which was also used in \cite{colloc2}, depicted in the lower left plot of Fig. \ref{pdes} for $\nu = 0.01/\pi$.

\subsection{Allen-Cahn equation}

Finally, the 1+1-dimensional nonlinear Allen-Cahn equation is expressed as:

\begin{equation*}
	\frac{\partial u}{\partial t} - D\frac{\partial^2u}{\partial x^2} + 5\left(u^3 - u\right) = 0,
\end{equation*}

\noindent where $\Omega = \left[0,1\right]\times\left[-1,1\right]$. For our experiments we chose $D = 0.001$, with the following boundary conditions:

\begin{align}
	u\left(t=0,x\right) &= x^2\cos\left(\pi x\right), \nonumber \\
	u\left(t,x=-1\right) &= u\left(t,x=1\right) = -1. \nonumber
\end{align}

\noindent Similar to Burgers' equation, this PDE does not have an analytical solution. Again, we used the reference solution from \cite{colloc2}, depicted in the lower right plot of Fig. \ref{pdes}.

\section{Adaptive State Transition in Function Learning} \label{appB}

\begin{figure}[H]
	\centering
	\includegraphics[width=\linewidth]{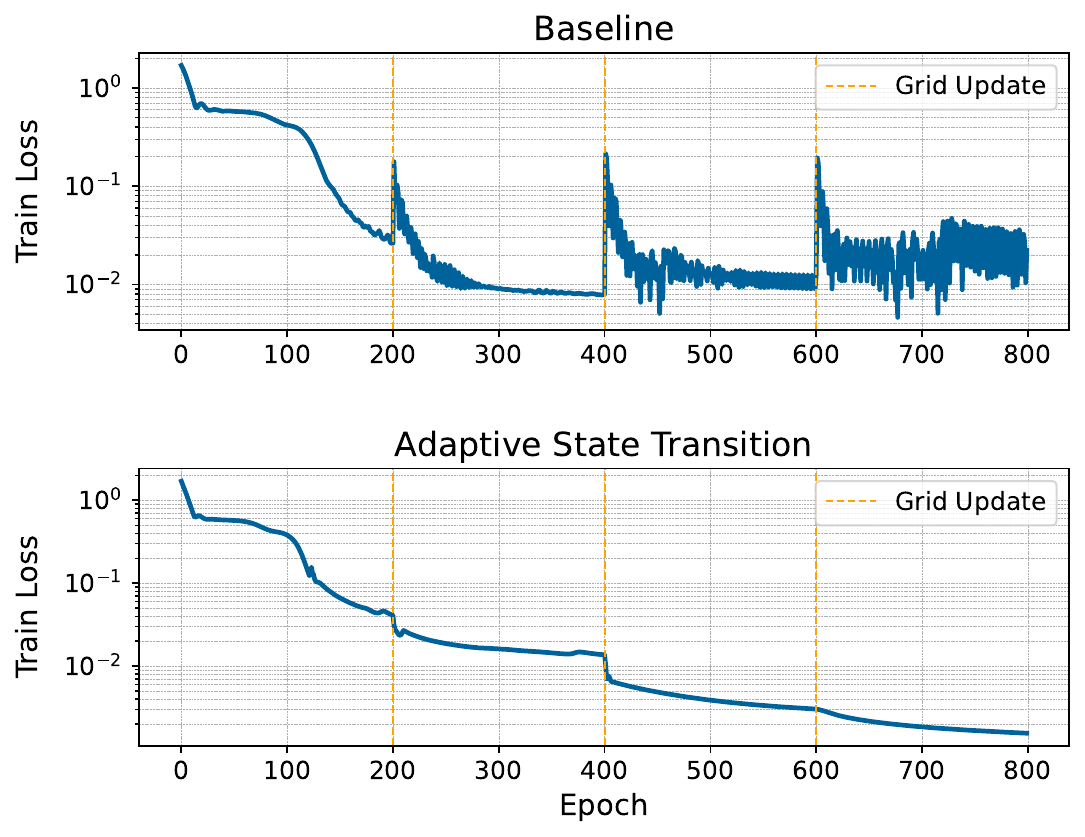}
	\caption{Training a KAN without (top) and with (bottom) the adaptive optimizer state transition technique.}
	\label{curve_state}
\end{figure}

In this Appendix, we present an application of the adaptive state transition technique introduced in Section \ref{sec:training}, for KANs that are trained to learn functions. This was also one of their main tasks in the work that introduced them \cite{pykan}. To this end, we trained two $[4,5,2,1]$ KANs with $g_e = 0.02$ and spline basis functions of order $k = 3$ to learn the multivariate function

\begin{align}
	f\left(x_1,x_2,x_3,x_4\right) = \exp&\left[\frac{1}{2}\sin\left(\pi x_1^2 + \pi x_2^2\right) + \right. \nonumber \\
	& \left. ~~\frac{1}{2}\sin\left(\pi x_3^2 + \pi x_4^2\right)\right], \nonumber 
\end{align}

\noindent by sampling 3000 points from it. The initial size of each KAN's grid was $G = 3$ and was extended to $G \to 6 \to 10 \to 24$ during epoch No. 200, 400 and 600, respectively. The baseline KAN was trained for a total of 800 epochs using a uniform learning rate of $0.02$ and no adaptive state transition for the optimizer. The other KAN was also trained for a total of 800 epochs, however with the adaptive state transition technique and a learning rate which started from $0.02$ and was scaled by a factor of 0.5, 0.2 and 0.5, during epoch No. 200, 400 and 600, respectively. The results for both KANs are depicted through the values of their loss functions during training in Fig. \ref{curve_state}.

\noindent This example showcases how the proposed technique reverses the peaks shown in the loss function's values during grid extensions: the sharp increases in the baseline's training are turned into sharp decreases for the training of the adaptive KAN. Additionally, the technique leads to overall better fits for the trained network, with the adaptive KAN achieving losses that are lower than those of the baseline KAN by approximately 1 order of magnitude.

\section*{Reproducibility}

The \texttt{jaxKAN} library is fully open-source and available through PyPI. The generated data, reference solutions and code used to produce the results of all experiments presented herein can be found in the library's GitHub repository, https://github.com/srigas/jaxKAN. All of our code is available under MIT license.

\section*{Acknowledgment}
S. R. thanks P. Angelopoulou for thoroughly reading the original draft to point out typos and inconsistencies.

\end{document}